# Mapping Process for the Task: Wikidata Statements to Text as Wikipedia Sentences

Thang Ta Hoang[a,b] (ORCID: 0000-0003-0321-5106), Alexander Gelbukh[a,*], Grigori Sidorov[a]
[a] *Computer Research Center, National Polytechnic Institute (IPN), CDMX, Mexico*
[b] *Faculty of Information Technology, Da Lat university, Da Lat, Viet Nam*
Emails: thangth@dlu.edu.vn, gelbukh@cic.ipn.mx, sidorov@cic.ipn.mx

**Abstract.** Acknowledged as one of the most successful online cooperative projects in human society, Wikipedia has obtained rapid growth in recent years, desires continuously to expand content, and disseminate knowledge values for everyone globally. The shortage of volunteers brings to Wikipedia many issues, including developing content for over 300 languages at the present. Therefore, the benefit that machines can automatically generate content to reduce human efforts on Wikipedia language projects could be considerable. In this paper, we propose our mapping process for the task of converting Wikidata statements to natural language text (WS2T) for Wikipedia projects at the sentence level. The main step is to organize statements, represented as a group of quadruples and triples, and then to map them to corresponding sentences in English Wikipedia. We evaluate the output corpus in various aspects: sentence structure analyzing, noise filtering, and relationships between sentence components based on word embeddings models. The results are helpful not only for the data-to-text generation task but also for other relevant works in the field.

## 1. Introduction

According to the Wikimedia movement, the mission of developing content in 2030 will go further what has done in the past to turn Wikimedia become "the essential infrastructure of the ecosystem of free knowledge"[1]. This mission inspires stakeholders to devote knowledge to Wikipedia language projects. To broaden efficiently a project with more content, one of the prerequisites is to have more volunteers. As a free and open encyclopedia, there is no obligation for a person to contribute to Wikipedia unless from his willingness. It is challenging for Wikipedia to entice a sufficient number of volunteers commensuration with its development phases. Thus, natural language generation (NLG) is an indispensable technology that Wikipedia must apply to create new natural language content for maintaining the pace of its development. Hundreds of new stub Wikipedia pages created by bots and upgraded then by human efforts make this strategy will soon become a popular perspective for Wikipedia projects.

Wikidata statements to text (WS2T) is a sub-task of data-to-text generation in NLG, both have results as natural language text that expresses sufficiently the input as structured data. WS2T is relatively new to the research community but data-to-text generation is not. In a few decades, data-to-text generation has gained a lot of achievements mostly in the media and healthcare industries [12], producing/synthesizing instant reports in various topics (weather [13, 14], soccer [15], financial [16], etc) or summarizing/ diagnosing medical information [17, 18, 19]. With the emergence of deep learning in the last decade, the task of data-to-text generation towards more automatic processes to convert structured data to natural language, sometimes ignoring the concern of transformations in traditional methods.

---

[1] https://meta.wikimedia.org/wiki/Strategy
[*] Corresponding author. E-mail: gelbukh@cic.ipn.mx.

In this paper, we research techniques and methods of how to map Wikidata statements to Wikipedia sentences in the mapping process and how to evaluate the output corpus as labeled sentences. Why do we choose Wikipedia and Wikidata as the two research objects in this paper? They are both sister projects and also are free online collaborations projects. The structured data of Wikidata is extracted from Wikipedia content and also be contributed by its volunteer community. We would like to do the inverse problem to convert structured data of Wikidata to Wikipedia to see how well these projects work together.

Founded in 2012, Wikidata is a pivotal bridge connecting Wikimedia projects together by interwiki link structure. Wikidata has accumulated from contributors massive, multilingual, and collaborative structured data [20] considered as a valuable source for Wiki-based projects and NLG applications. Wikidata statements can be represented as RDF triples, displays characteristics of an entity (item/property). In other words, structured data consists of features as Wikidata statements. We assume that Wikipedia content is a "valuable" reference source for any content generation system which reflects the style of writing, word usage, phraseology, and grammar structure of millions of volunteers. Thus, the translation task is expected to mimic Wikipedia as best as possible in outcome sentences.

Our research scope is limited at sentence level with single sentences but not summaries/first sentences in English Wikipedia pages. To reduce the mapping complication, we only use sentences that contain only a subject and possibly a verb. An RDF triple is commonly translated to a very short sentence but it may not be popular usage in our written language. For example, *isa("Volga", "river")* can be generated into a simple sentence, "*Volga is a river*". To generate a longer and more practical sentence, a set of RDF triples is used. When combining two triples, *location("Volga", "Russia")* and *length("Volga", "3530 km")* with the previous triple, a better sentence could be "*Volga is a river with a length of 3530 km, located in Russia*". When using a quad, a triple with extended properties, we can generate a sentence with acceptable length to use in Wikipedia pages. Of course, from a set of quad, we can be able to produce a high-quality sentence or even a paragraph. However, we prefer more to work with a single quad to learn characteristics between it and Wikipedia sentences.

We evaluate our corpus, containing 18510 sentences in two aspects: the mapping process and the quality of labeled sentences in the output. The mapping process is assessed by applying entity linking methods to score data matching and type matching for each item in a quad and each component of a matched sentence corresponding to this quad. We also do some basic statistics on labeled sentences. Furthermore, we extract candidate sentences by our definition and apply clustering algorithms to filter outliers or noises. We mainly depend on Word2vec models to assess the word relatedness (semantic similarity) between labeled sentences and their corresponding quads by statement context to see some relationships. Statement context takes labels and aliases from every quad item to create a context set, used to measure word vector distances between labeled sentences and these quads. We realize that a labeled sentence is considered high-quality only when it contains all words (except stop words) related to the context set. From that, this relatedness can be ranked on a descending score for every labeled sentence.

Beyond this section, our paper consists of other sections in the following order. Recent works on the task of generating natural language content from structured data related to Wikipedia and Wikidata are described in Section 2. We will introduce our understanding of Wikidata structure and its statement classification in Section 3. Section 4 defines methods and formulas of the mapping process from Wikidata statements to Wikipedia sentences. Section 5 presents our techniques and algorithms in choosing several typical Wikipedia properties for the data collection process. The evaluation of the output corpus, including basic statistics, noise filtering, and relationships between Wikidata and labeled sentences will be presented in Section 6. Finally, we make some conclusions and declare our future work in Section 7.

## 2. Literature Review

Most of the related papers work on generating brief summaries or first few sentences of Wikipedia pages, where may contain a high semantic density, especially *isa* relations supposed to attain better outcome performance. In contrast, it is less popular to translate structured data to a Wikipedia page (stub pages) [10, 11, 26] or to sentences, which do not belong to the first paragraph as our research. The input, structured data, taken from available data sources (DBPedia [2, 4, 21, 23], Wikidata [3, 4, 7, 21, 23]), is organized in different formats, from RDF triples [2, 3,

4, 9], tables [5, 7, 8] (Wikipedia infoboxes, slot-value pairs), to knowledge graphs [22, 24] built from RDF triples. In our paper, we refer from the works [1, 27, 28, 29, 30] to format Wikidata statements in a set of quads and triples which is possible to transform to RDF triples for creating knowledge graphs.

There are two significant approaches, sentence templates or placeholders and neural networks (CNN, Seq2Seq, GCNs) for yielding Wikipedia sentences from structured data. Duman and Klein applied LOD-DEF [2], a proof-of-concept system that learns sentence templates and statistical document planning to yield natural language from RDF graphs. Authors proposed to prune redundant words if not fit to the templates but this action somehow ignores helpful information which may recognize in the extra matching process. Sauper and Barzilay leveraged Internet sources and human-authored texts to infer templates with beneficial results when putting structural information in the content selection task [6].

To reduce the dependence of templates, vector space models and neural networks are spotlights for overcoming the subjective of selecting input data features. Kaffee et. al. built a neural encoder-decoder architecture to extend ArticlePlaceholders for creating multilingual summaries for Wikipedia pages with better performance compared to MT and template-based methods [3, 53]. Vougiouklis et. al. used a set of triples, converted to a fixed-dimension vector, then put it to an end-to-end trainable architecture to receive an encoded vector, which is taken to translate to an output summary [4]. Work on the biology domain, Lebret et. al. introduced a conditional-based neural model, which translates first sentences from a large-scale dataset, performing an outstanding result compared to Templated Kneser-Ney language model [5]. Their dataset, WIKIBIO contains over 700k articles with infoboxes taken from English Wikipedia. Later on, Liu et. al. reused WIKIBIO to work with the task of table-to-text generation, applied a novel structure-aware seq2seq architecture, using field-gating encoder in the encoding phase and dual attention mechanism in the decoding phase [8]. Chisholm et. al. trained a CNN sequence-to-sequence model where focuses on input facts to be sure that they appear in the output, biographical sentences [7]. Also, applying a sequence-to-sequence model, Zhu et. al. realized that loosely sentences tend to cause by maximum likelihood estimation in trained language models. Their solution is to use the inverse Kullback-Leibler (KL) divergence [9], calculating the distribution difference between the reference and outcome sentences in a novel Triple-to-Text (T2T) framework.

Different from sequence-to-sequence methods, Mar-cheggiani and Perez-Beltrachini introduced a structured data encoder based on graph convolutional networks (GCNs), which obtains advantages from exploiting input structure and respecting to sequential encoding. Trisedya et. al. proposed a graph-based triple encoder GTR-LSTM which captures triple information as much as possible, indicated better results compared to some baseline models [24]. A decoder-only architecture presented by J. Liu et. al, is used to produce long and very long sequences, has many potentials to generate fluent, coherent content [26]. We consider this research as one of our future works for generating stub pages for Wikipedia.

There is also a third approach that relies on the combination between sentence templates and neural networks. Luo et. al. proposed [48] a text stitch model which overcomes the drawbacks of both neural networks and template-based systems. This model allows us to control the input information to maintain the fidelity of the output, and to reduce the human efforts in building templates. In another paper, Wiseman et. al constructed a neural generation model based on an HSMM decoder [50] to discover hidden templates for improving the generation diversity and creating interpretable states.

We refer some ideas from [21, 31] to build methods for the content selection and matching steps for each item in Wikidata quads to Wikipedia sentences. Word2vec models [25] will be applied to rank the relatedness of labeled sentences to the Wikidata statement context and to search for the relationships between these sentences against Wikidata properties or qualifiers.

## 3. Wikidata repository

In Wikidata, structured data is mainly organized into two multilingual objects: item (entity) and property. Every item or property consists of an identifier (primary key), a label, a description, statements, and aliases, i.e, alternative names of its label. Both item and property identifiers have a similar format, starting with letter "*Q*" (item) or "*P*" (property) and following by a positive integer suffix. For example in Figure 2, *Q1372810* refers to an item with label "*Simone Loria*" and *P54* refers to a property with label "*member of sports team*". We will use Figure 2 and Figure 3 as examples to describe data structure of a typical item in Wikidata, which readers can refer for many other sections of this paper.

An item can comprise many interwiki links, connecting to other content entities (pages, categories, templates, etc) in Wikimedia projects and Wiki-based systems together; and extra identifiers, linking this item to other data resources. Each interwiki link denotes a language representation which appears in the 3-column tabular form: a language syntax (e.g. *en* denoted for English Wikipedia), an item label and a URL link to that language. Meanwhile, a property includes a data type, and constraints, defining data rules for that property value.

*3.1. Data types in Wikidata*

In this section, we mention data types related relatively close to our work because Wikidata captures a wide range of different data types[2] in its data structure. In Wikidata, "*wikibase-item*" is a data type that can apply to both items and properties, representing a connection ability to another item. Likewise, "*wikibase-property*" only used for properties, shows a link between a property with another property. Properties normally take all of data types, some as presenting in Table 3.

Table 1

Some typical data types used in Wikidata.

| Data type | Description | Object applied |
|---|---|---|
| wikibase-item | internal link to another item | Item Property |
| wikibase-property | internal link to a property | Property |
| string | sequence of characters, numbers and symbols | Property |
| quantity | a decimal number with a range (upper and lower bounds) and a unit of measurement | Property |
| time | a date in Gregorian or Julian calendar | Property |
| URL | a reference to a web or file resource | Property |
| globe-coordinate | a geographical position given as a latitude-longitude pair | Property |

Depend on internal discussions in Wikidata, a property bound to which a unique data type is used for which statements. Any modifications of structure data will not be changed until gaining the community consensus. In some aspects, we can say that data types have a certain impact on the behavior of statements because of data values they take in their properties. A statement may be different from another one by properties and data types it uses.

*3.2. Term usage*

When reviewing related works, we realize that there are a few variant terms referring to elements of data structure, mainly statements in Wikidata. So we would like to present all terms and their synonyms, referring from Wikidata glossary[3] in Table 2. Our purpose is to wipe out the ambiguation of readers, help them to follow and to understand our term usage.

Table 2

Wikidata terms and its variants.

| Wikidata | Other usage | Description |
|---|---|---|
| claim | statement | a single claim belongs to the statement group |
| item | subject* | a real-world object, concept, or event |
| item identifier | primary key | an item's identifier |
| property identifier | primary key | an property's identifier |
| property | predicate* | a property of an item |
| value | object*, property value | a property value |
| qualifier | extend property, propery term | a property to extend the context of a claim |
| alias | alternative names, term | a group of alternative names of the label |
| label | term | a item's label |
| description | term | a brief definition of an item |
| * Our preferable usage in the mapping process | | |

*3.3. Wikidata statements*

Although statements are elements defined for both items and properties, we only focus on those belong to items. An item contains statements called statement group, which are treated as features that show characteristics for that item. Different items may take different statements and items clustering in the same group may share some similar statements.

A typical statement includes only one property, which is able to take multiple values, which we prefer to call "objects" in Table 2. Each value has a rank, references and may contain a set of qualifiers (property terms) with their values. A value rank (object rank) is used to show its order compared to other values in the same property. References are sources (web links, wiki projects, corpus, etc) where a value and its qualifiers are collected. Qualifiers provide the external context for a statement while a traditional statement *(item, property, value)* can not offer.

---

[2] https://www.wikidata.org/wiki/Help:Data_type

[3] https://www.wikidata.org/wiki/Wikidata:Glossary

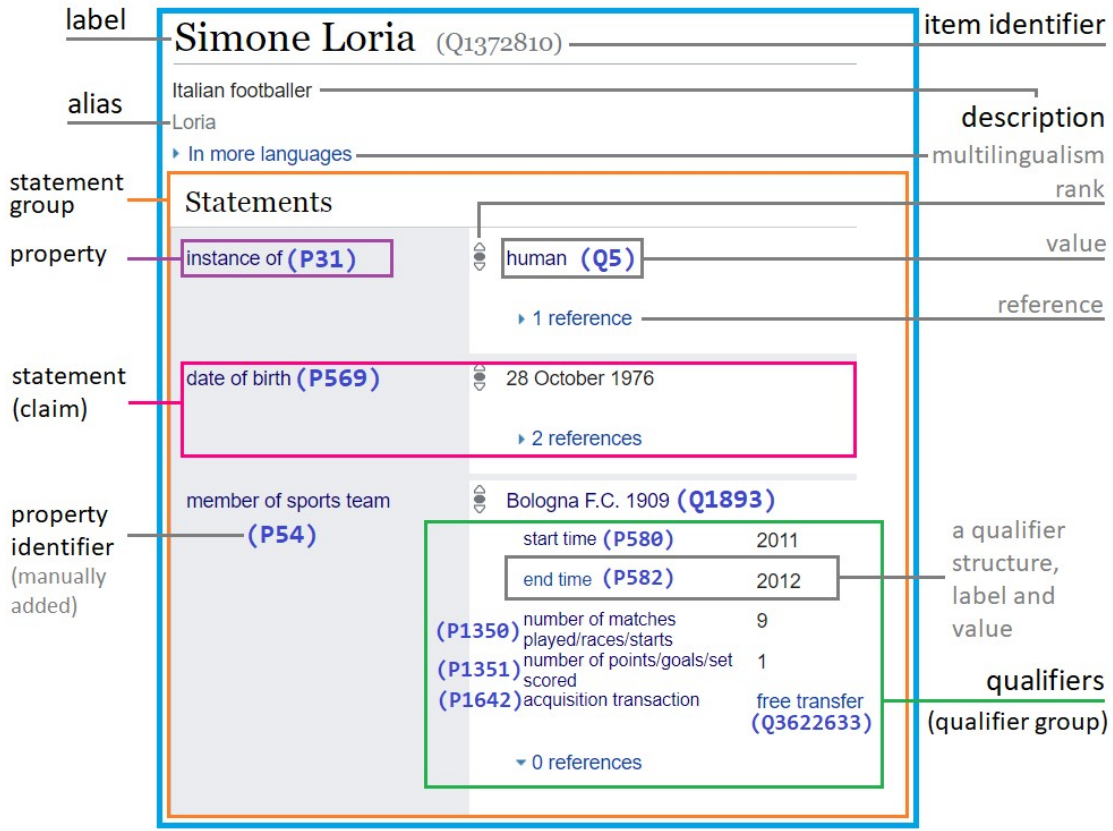

Fig. 1. Data structure of item Q137280 in Wikidata.

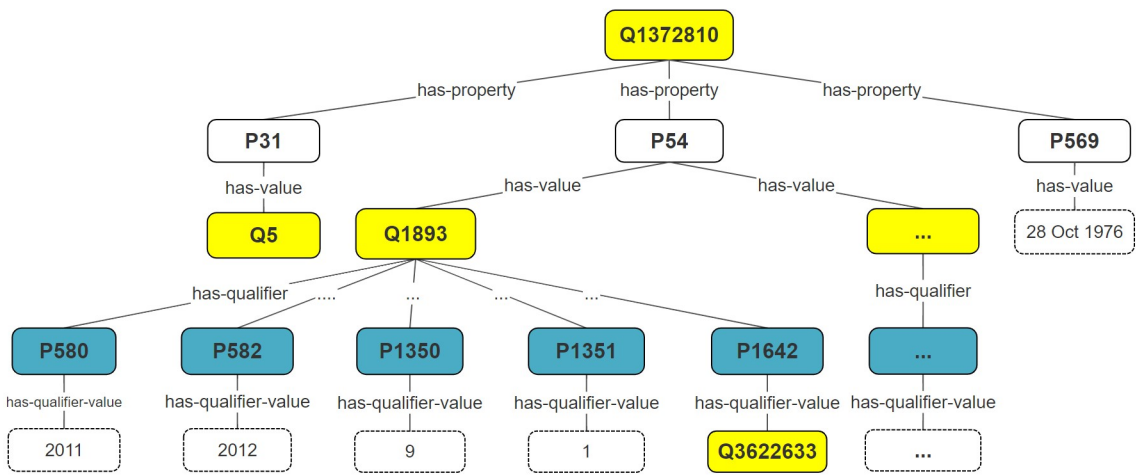

Fig. 2. Statemements of item Q137280 organized as tree structure.

In Figure 1, item *Q1372810* has three statements corresponding to three properties, *P31, P569* and *P54*. We can represent these statements as tuples:
- *(Q1372810, P31, Q5)*,
- *(Q1372810, P569, "28 October 1976")*, and
- *(Q1372810, P54, Q1893, ((P580, 2011), (P582, 2012), (P1350,9),..., (P1642, Q3622633)))*.

If temporarily take the rank and references out, a statement is viewed as a multi-node tree with a root node is its property, whose values are children. Each value also has children as its qualifiers. In a similar way, we can form a tree for the statement group with the item as a root node. Figure 2 is an example of how to represent a multi-node tree for the statement group.

From the tree structure, we notice an interesting characteristic of statements. If a node (in the yellow or light background) is an item, it can extend the tree span by adding its child nodes. Otherwise, if a node is a property (in pacific blue or dark background), the tree has no scalable ability.

### 3.4. Statement classification

The difference in using property and qualifier values urge us to classify statements into three different types. This task is to aim at two objectives: to understand how to transform Wikidata statements to RDF triples and to make clear how to deserialize and organize Wikidata's structured data in the format of XML or JSON.

#### 3.4.1. Wikidata statement type 1

Wikidata statement type 1 (WST-1) is a statement containing an item, a property, property values, and without qualifiers. The data type of the property values (object) is not *wikibase-item*.

Normally, WST-1 has only one property value. If there are multiple property values in a statement, we can separate it into smaller statements by these values. The phenomenon "a property/qualifier with multiple values" is called heterogeneity, which usually happens in all Wikidata statements because they retrieved from Wikipedia, whose the content contributions come from so many different volunteers. The heterogeneity happens not only for this statement type but also for all other statements in Wikidata, but we prefer to take this problem out of the mapping process.

For example, in Figure 1, we have statement *(Q1372810, P569, "28 October 1976")* is a WST-1 and its propery value is a datetime, *"28 October 1976"*. If this statement has another property value (object) supposed as *"28 October 1977"*, we can represent it as one statement, *(Q1372810, P569, ("28 October 1976"," 28 October 1977"))* or two independent statements, *(Q1372810, P569, "28 October 1976")* and *(Q1372810, P569, "28 October 1977")*.

#### 3.4.2. Wikidata statement type 2

Like almost of WST-1, Wikidata statement type 2 (WST-2) also has no qualifiers but the data type of its property values is *wikibase-item*. For example, in Figure 1, we have statement *(Q1372810, P31, Q5)* is a WST-2 with its propery value is an item, *Q5*. For a statement with multiple objects, we will also split it into smaller statements. Furthermore, the property value in this statement can be scalable in connecting to other items or its properties.

#### 3.4.3. Wikidata statement type 3

Wikidata statement type 3 (WST-3) is a statement having all four elements, *(item, property, property value, qualifier)*. It is the richest structured data in Wikidata statements, which should be exploited to get more semantic sources.

At first, we presume that WST-3 should has an item, a property, property values (objects), qualifiers (for each property value) and qualifier values. However, this makes WST-3 structure too bulky, so we decide that WST-3 will contain an item, a property, a property value, qualifiers, and qualifier values.

WST-3 can continue to decay itself into smaller statements, each of which has an item, a property, a property value, a qualifier, and a qualifer value. Here, it is akin to the difference between WST-1 and WST-2 on the property value, we also care about the qualifer value to define two sub-types of WST-3.
- WST-3a: a WST-3 with data type of the qualifier value is not a *wikibase-item*.
- WST-3b: a WST-3 with data type of the qualifier value is a *wikibase-item*.

Besides, we treat the heterogeneity (mutiple qualifiers/qualifier values) for WST-3 statements exactly the same as WST-1 and WST-2 statements.

In Figure 1, we have a WST-3, that is *(Q1372810, P54, Q1893, ((P580, 2011), (P582, 2012), (P1350,9),..., (P1642, Q3622633)))*. We can split it up to these statements:

– *(Q1372810, P54, Q1893, (P580, 2011))*,
– *(Q1372810, P54, Q1893, (P582, 2012))*,
– *(Q1372810, P54, Q1893, (P1350, 9))*,
– *(Q1372810, P54, Q1893, (P1351, 1))*,
– *(Q1372810, P54, Q1893, (P1642, Q3622633))*.

The first four statements are WST-3a and the last one is WST-3b.

*3.5. Wikidata structure encoding*

Statements must be constructed in a clear, concise structure and have connections with other semantic structural types, such as RDF triples. To encode Wikidata as RDF, there are various ways [1, 28, 32, 33], from standard reification, n-ary relations, singleton properties, named graphs, to property graphs. We are inspired by Hernández et. al [1] to present Wikidata statements in two tables, QUAD and TRIPLE but in a slightly different way.

*3.5.1. QUAD table*

This table will hold a set of quads in the form *(s, p, o, trip_id:q)*, which corresponds to *(subject, predicate, object, trip_id:qualifier)*. This form will also be *(item, property, value, trip_id:qualifier)* as mentioned about our term usage in Table 2. The syntax *trip_id:q* means *trip_id* indicates to an identifier, generated for a tuple *(s, p, o)* in a relationship with qualifier *q*. Note that, we only store identifiers (item or property identifers) for elements in every quad.

WST-3a and WST3-b statements are proper objects to be kept in the QUAD table. Table 3 demonstrates how to store WST-3 statements from Figure 1.

Table 3
Store statements from Figure 1 as QUAD table with the form (s, p, o, trip_id:q) in every row.

| s | p | o | trip_id:q |
|---|---|---|---|
| Q1372810 | P54 | Q1893 | trip_1: P580 |
| Q1372810 | P54 | Q1893 | trip_1: P582 |
| Q1372810 | P54 | Q1893 | trip_1: P1350 |
| Q1372810 | P54 | Q1893 | trip_1: P1351 |
| Q1372810 | P54 | Q1893 | trip_1: P1642 |

An identifier value is established in triple *(Q1372810, P54, Q1893)* as *trip_1*, which marks the relationship with its qualifiers. Instead of holding qualifier values in this table, we keep them in TRIPLE table.

To avoid the repetition of information (the same triple for every qualifier), it is possible to optimize the QUAD table follow the form *(s, p, o, trip_id:QL)* where *QL* is a set of qualifiers. Therefore, Table 3 just needs to store a single tuple, *(Q1372810, P54, Q1893, trip_1:"P580, P582, P1350, P1351, P1642")*.

*3.5.2. TRIPLE table*

TRIPLE is a table containing RDF triples, converted from WST-1 and WST-2 statements as well as item terms (label, description, and alias). For every row, we define tuple *(x, y, z)*, which is also considered as *(subject, predicate, object)* or *(item, property, value)*.

Extended information constructed as triples from each element in QUAD will be stored in TRIPLE. Even, a triple element itself may trigger to store new triples, so data expansion can occur to a very large scale in TRIPLE. For example, we will extract and store structured data from Figure 1 as triples in Table 4.

Table 4
Use TRIPLE table to store statements and structured data in the tuple form (x, y, z) in Figure 1.

| x | y | z |
|---|---|---|
| Q1372810 | :label | "Simone Loria" |
| Q1372810 | :description | "Italian footballer" |
| Q1372810 | :alias | "Loria" |
| Q1372810 | P31 | Q5 |
| P31 | :label | "instance of" |
| Q5 | :label | "human" |
| Q1372810 | P569 | "23 October 1967":datetime |
| P569 | :label | "date of birth" |
| triple_1:P580 | :value | "2011":datetime |
| P580 | :label | "start time" |
| P54 | :label | "member of sports team" |
| Q1893 | :label | "Bologna F.C. 1909" |
| … | … | … |

Labels, descriptions, and aliases should be archived at first because they provide important information in the mapping process to recognize which item or property in the semantic aspect. In Table 4, these values in turn for *Q1372810* are "*Simone Loria*", "*Italian footballer*", and "*Loria*". In another example, it is arduous to perceive the information of WST-1 statement *(Q1372810, P31, Q5)*, thus aside from *Q1372810*, *P31* and *Q5* should be clarified by adding two triples, *(P31, :label, "instance of")* and *(Q5, :label, "human")* to the storage. After that, we can easily sense the semantic meaning of label triple *("Simone Loria", "instance of", "human")*.

A qualifier value in TRIPLE is linked to a quad in QUAD through syntax *triple_id:q* to determine a connection between them. Look up to both Table 3 and Table 4, two tuples, *(Q1372810, P54, Q1893, trip_1:P580)* and *(trip_1:P580, :value,*

*"2011":datetime)* can be associated each other by *trip_1:P580* in order to get the qualifier value of *P580* is *"2011":datetime*.

## 4. Mapping Wikidata statements to Wikipedia sentences

### 4.1. A thorough example

In this paper, the most important task is to select appropriate sentences from mapping each row in the QUAD table to English Wikipedia content. Before introducing our methodology, we would like to bring up a meticulous example to support amateur readers to catch up with what we will execute.

Figure 3 shows how to map successfully a quad to a sentence in Wikipedia and create a labeled sentence from matching pairs.

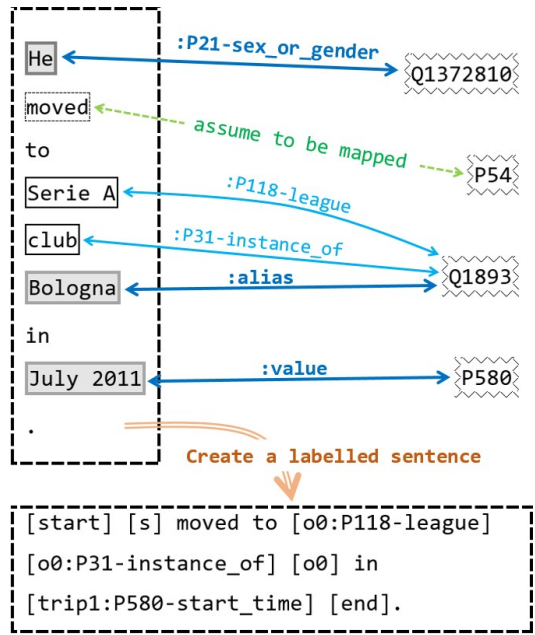

Fig. 3. The sketch of how quad (Q1372810, P54, Q1893, P580) is mapped to a sentence. The bold lines are the main mappings, the thinner lines are the extra matchings and the dashed line is assumed that there has a connection. From the matching pairs, the sentence will be transformed into a labeled sentence.

Different from the theory, which required to map all elements of a quad to a sentence's components, the experiment only works with elements: *s*, *o*, and *q* (*subject, object, qualifier*). Element *p* (predicate or property) is a description of the relationship between *s* and *o*. In a certain sentence, *p* should align to a verb or a verb phrase. However, *p* takes the form of a set of phrases (mostly nouns/noun phrases) in Wikidata. So it is rare to map successfully from a verb to a set of noun phrases. We thus avoid the mapping for *p* (the predicate matching) but will evaluate it later in Section 6.

The scenario of an English Wikipedia page linked to an item in Wikidata via an interwiki link, is also applied for other languages. In another saying, an item in Wikidata possibly is associated with multiple page, each of which belongs to a unique language project. In Figure 1, the content of the page titled "*Mario Kirev*", organized as a set of sentences is interlinked to Wikidata item "*Q1372810*", whose structured data is converted into the tabular form in Table 5. Note that we always execute the mapping between a page and an item having an interwiki link to assume that the success possibility is highest.

Table 5

Encode Wikidata statements for the case of Figure 3 by a set of quads and triples, splitting into two tables.

**QUAD**

|   | s | p | o | triple_id:q |
|---|---|---|---|---|
| 1 | Q1372810 | P54 | Q1893 | trip1:P580 |

**TRIPLE**

|    | x | y | z |
|----|---|---|---|
| 1  | Q1372810 | :label | "Simone Loria" |
| 2  | Q1372810 | :description | "Italian footballer" |
| 3  | Q1372810 | P735 | Q18067255 |
| 4  | Q18067255 | :label | "Simone" |
| 5  | Q1372810 | P21 | Q6581097 |
| 6  | Q6581097 | :label | "male" |
| 7  | P753 | :label | "give name" |
| 8  | P21 | :label | "sex or gender" |
| 9  | P54 | :label | "member of sports team" |
| 10 | P54 | :description | "sports teams" |
| 11 | P54 | :aliases | "member of team" |
| 12 | Q1893 | :label | "Bologna F.C. 1909" |
| 13 | Q1893 | :description | "association football club" |
| 14 | Q1893 | :alias | "Bologna, Bologna Foot…" |
| 15 | P580 | :label | "start time" |
| 16 | P580 | :description | "time an item begins to..." |
| 17 | P580 | :alias | "from, starting, began,..." |
| 18 | trip1:P580 | :value | "+2011-01-01T00:00:00Z":datetime |
| 19 | Q1893 | P118 | Q1584 |
| 20 | P118 | :label | "league" |
| 21 | Q1584 | :label | "Serie A" |
| 22 | Q1893 | P31 | Q476028 |
| 23 | P31 | :label | "instance of" |
| 25 | Q476028 | :label | "association football club" |
| 26 | Q476028 | :description | "sports club devoted to…" |
| 27 | Q476028 | :alias | "football club, soccer club" |
| 28 | Q1372810 | :alias | "Loria" |

Here is the explanation for the mapping process in Figure 3. From Table 5, set *{"Simone Loria", "Simone", "he","Loria"}* is extracted for the subject matching. Triple (row: 1) allows us take value "*Simone Loria*". From triples (rows: 3, 4, 7), we have value "*Simone*" and infer from triples (rows: 5, 6, 8) for value "*he*". From a triple (row: 28), we have value "*Loria*". Then, the result of subject matching is pair *(Q1372810, "he")*. Next, set *{"Bologna F.C. 1909", "Bologna", "Bologna Foot…",…}* is extracted from triples (rows: 12, 14) to map to word "*Bologna*" in the sentence, making object matching pair *(Q1893, "Bologna")*. We do a similar thing to qualifier matching pair *(P580, "July 2011")*, inferring the alignment between set *{"+2011-01-01T00:00:00Z"}* extracted from triple (rows: 15, 18) and the sentence text. The syntax "*trip1:P580*" means qualifier *P580* is in a relationship with triple *(Q1372810, P54, Q1893)*. This is to be sure that we can get the qualifier value of *P580* exactly by its triple relationship. Remember that *P580* could be a qualifier of another triple and its qualifier value with this triple may be different. The pairs, *(Q1893:P31:Q476028:label, "club")* and *(Q1893:P118:Q1584:label, "Serie A")* are the extra matchings that respect to object *Q1893*, comparing sentence phrases with triples (rows: 19 – 26). We keep these matching pairs in short forms as *(Q1893:P31, "club")* and *(Q1893:P118, "Serie A")*.

Table 6

Some predefined rules for labeling result pairs to the example taken from Figure 3.

| Label | Description | Figure 3 |
|---|---|---|
| [start] | A syntax to start a sentence | |
| [end] | A syntax to end a sentence, before a true full stop mark (.) | |
| [s] | The subject of a sentence | He --- [s] |
| [o$_0$] | the first object of a sentence | Bologna --- [o0] |
| [o$_1$] | the second object of a sentence | |
| … | … | |
| [trip1:Pxxx-yyy] | Qualifier Pxxx with label "yyy" for a triple (s, p, o) | July 2011 --- [trip1:P580-start_time] * <br><br> trip1 = (Q1372810, P54, Q1893) |
| [o$_n$:Pxxx-yyy] | The extra matching for Pxxx with label "yyy" respects to o$_n$. | Serie A --- [o0:P118-league] |

* Syntax [o0:P580-qualifier] is used in the experiment.

The last matching step is to label result pairs to the input sentence that complies with some predefined rules in Table 6. The labeled sentence is viewed as a template applied for generating sentences from similar quads.

For the case (*) in Table 2, we can use *[o0:P580-qualifier]* instead of *[trip1:P580-start_time]* in the experiment because there are fewer conflicts between qualifiers for the mapping at sentence level in separate pages. Furthermore, this change is more comfortable to identify which qualifier belongs to which object when dealing with multiple objects and qualifiers.

### 4.2. Quad structure definition

In this section, we will define the structure of a quad and its components used in the mapping process. A quad is a WST-3 statement in the form *(item, property, property value, (qualifiers, qualifier values))* or *(subject, predicate, object, (qualifiers, qualifier values))* or even in a short form *(s, p, o, Q)*, where *Q* is a set of qualifier-value pairs. However, to maximize the matching ability, we use a bulky WST-3 statement as mentioned in Section 3.4.3. In that case, the structure of quad *T* should be:

$$T = (s, p, O_n, Q_m \mid \forall n, m \in N_1) \quad (1)$$

In (1), *s* and *p* must exist. $O_n$ is a set of objects (values) with *n* elements. Similarly, $Q_m$ is a set of qualifiers with *m* elements, each of which is in the form *(object, qualifier, qualifier value)* or *(o, q, qv)* to know which object it belongs to. $O_n$ and $Q_m$ must carry at least one element.

Quad *T* can decay to WST-3 (WST3-a, WST3-b) statements to store in *QUAD* and *TRIPLE* tables in Section 3.5.1. Besides, quad *T* is a child case of a bigger quad with multiple predicates, should be in the form *(s, $P_k$, $O_n$, $Q_m$)*. This new quad indeed covers all statements of an item in Wikidata.

Set $O_n$ is a set of values of *p* and it is not empty. We can formulate this set as:

$$O_n = \{o_1, o_2, o_3, ..., o_n \mid n \in N_1\} \quad (2)$$

Every $o_i$ contains its qualifier sets, called $Y(o_i)$. For example, in Figure 2, object *Q1893* contains qualifiers $Y(o_{Q1893})$ = *{(Q1893, P580,"2011"), (Q1893, P582,"2012"), (Q1893, P1350,9), (Q1893, P1351,1),*

*(Q1893,P1642,Q3633633)}*. We define set $Y(o_i)$ including:

$$Y(o_i) = \{(o_i,q_1,v_1),(o_i,q_2,v_2),...,(o_i,q_x,v_x) \quad (3)$$
$$| x \in N_0, Y(o_i) \subseteq Q_m\}$$

with each $q_i$ belongs to set $Q_m$ in (1) and $x$ is the number of elements of $Y(o_i)$. We expect $x$ is equal and larger 1 in the experiment but it could take the value of zero. We do not deal with the heterogeneous case, which a qualifier can have multiple values, making the mapping more complicated.

According to (1), note that the minimum number of elements in set $Q_m$ is 1. Because a qualifier always belongs to an unique object, this leads to set $O_n$ always has at least one element, reaffirming the condition of $O_n$ in (1). From (2, 3), set $Q_m$ can be defined as:

$$Q_m = \{Y(o_1) \sqcup Y(o_2) \sqcup ... \sqcup Y(o_n) | \forall n \in N_1\} \quad (4a)$$

The number of qualifier elements of set $Q_m$ depends on the number of elements of its subsets (such as $Y(o_i)$) in (3), so $m$ can take the formula:

$$m = x_1 + x_2 + ... + x_y = \sum_{i=1}^{y} x_i \quad (4b)$$

with $x_i$ is the number of elements of $Y(o_i)$.

Now, to map a quad $T_i$ to a set of sentences $S$, we form the formula:

$$M(T_i, S) = T_i \rightarrow S_l, l \in N_0 \quad (5)$$

In (5), $S_l$ is a set of $l$ outcome sentences. When the mapping fails, $l$ will equal zero, or there is no sentence is mapped. Quad $T_i$ is able to map to many sentences but a sentence $s_j$ in S must belong to a unique quad $T_i$. Otherwise, if $s_j$ is aligned with many quads, we will drop this mapping.

*4.3. Mapping methods*

The mapping priority for a given quad $T_i$ is by this order: subject (*s*), object ($O_n$), qualifiers ($Q_m$), extra matchings with a set of triples extented from matched objects $O_t$, and predicate (*p*). So the mapping process is divided into five sub-processes:
– subject matching (mandatory),
– object matching (mandatory),
– qualifier matching (mandatory),
– extra matching (optional),
– and predicate matching (mandatory).

Beyond the predicate matching, other matchings are mandatory to do, which means the mapping will be discarded if missing the "correct" results from one of those. As mentioned in Section 4.1, there is an issue in the predicate matching, mostly failures when mapping between two sets, verbs and noun phrases. We will evaluate this step after the mapping process has been done because we want to know how many verbs so far can fit the input quads. However, in this section, we still love to introduce a solution of how we can handle the discrepancy between verbs and noun phrases (terms). The extra matching represents additional mappings that we can learn to produce longer sentences (templates) and how to connect to other triples. This may turn to a problem of forming knowledge graphs that can use to translate Wikidata statements to a group of sentences or paragraphs.

In each step above, we apply mainly value matching for noun phrases as the key method to map statements to sentences. Given phrase *c* and phrase *d*, there are several ways to do the matching for these two phrases:

– *Exact noun-phrase matching*: compare two phrases with exact string, regardless of sensitive and non-sensitive cases.
– *Partial noun-phrase matching*: compare parts of phrase *c* to phrase *d* or vice versa. We probably use word root $wr_c$ to compare to *d* or word root $wr_d$ to compare to *c*. For example in Figure 3 and Table 5, word "*club*" is mapped to the label of *P118:Q476028*, "*association football club*". This is a sub-case of the hypernym matching.
– *Hypernym matching*: Instead of comparing two phrases to each other, we compare a phrase to hypernyms of the other phrase or vice versa. This method is applied only when phrases are contained in a hierarchy system (ontology), such as WordNet. In Wikidata, we can take the label of property P31 (instance of) as a phrase value for the comparison.
– *Synonym matching*: compare a phrase with synonyms of the other phrase or vice versa.

We only apply the first two matchings whose deployment cost is relatively cheap but effectively in the experiment.

In Wikidata, aliases can substitute for the label so for each item or property, we define *G*, a union set of those two, holding mainly nouns or noun phrases.

Our purpose is to enlarge the matching possibility for each element in a quad. We define set $G$ as:

$$G = \ell \cup Al = \{g_1, g_2, ..., g_n \mid n \in N_1\} \quad (6a)$$

with $\ell$ is the label and $Al$ is the set of aliases. Every element $g_i$ of $G$ will be in English because the mapping process works with English Wikipedia. In Figure 3 and Table 5, set $G$ for $Q1372810$ is $\{$'Simone Loria', 'Loria'$\}$.

For a certain sentence, we can easily apply syntactic analysis methods to retrieve three sets: subjects, verbs, and entities from its structure. Entities are nouns or noun phrases collected by unifying named entities from named entity recognition (NER), noun chunks, and noun chunk roots (root words) in the parsing process. We can describe these disjoint sets as $U, V$ and $E$ in turn for subjects, verbs and entities. Each element in these sets is in the form *(term value, start position, end position, type, root word)* or *(tv, sp, ep, ty, rw)*. We define this form as the formula:

$$term_i = (tv_i, sp_i, ep_i, ty_i, rw_i) \left| \begin{array}{l} ep_i >= sp_i \\ ep_i, sp_i \in N_0 \end{array} \right) \quad (6b)$$

The end position must be larger or equal the start position and both of them are non-negative integers. We may use $tv_i$ to refer to the term value of a certain term in the matchings. For example, we extract and organize terms from the sentence in Figure 3 into three sets:
- $U$ = {('He', 0, 0, 'PRON', 'He')}
- $V$ = {('moved', 1, 1, 'ROOT', 'moved')}, and
- $E$ = { ('Serie A club', 3, 5, 'ORG', 'club'),
  ('Bologna', 6, 6, 'FAC', 'Bologna'),
  ('July 2011', 8, 9, 'DATE', '2011'),
  ('July', 8, 8, 'NOUN CHUNK', 'July'),
  ('Serie A', 3, 4, 'NOUN CHUNK', 'A') ,
  ('club', 5, 5, 'NOUN CHUNK','club'),
  ('A club', 4, 5, 'NOUN CHUNK','club'),
  ('Serie', 3, 3, 'NOUN CHUNK','Serie'),
  ('A', 4, 4, 'NOUN CHUNK','A')}

So now is for matchings. From (1) and (6a), we can split the mapping $M$ between quad $T_i$ and sentence $s_j$, denoted as $M(T_i, s_j)$ into five formulas corresponding to five matching steps above:

$$M_s(s,U) = G_s \rightarrow U \quad (7a)$$

$$M_O(O_n, E) = G_{O_n} \rightarrow E \quad (7b)$$

$$M_Q(Q_x, E_y) = G_{Q_x} \rightarrow E_y \text{ with } \begin{cases} Q_x \subseteq Q_m \\ E_y \subseteq E \end{cases} \quad (7c)$$

$$M_{Ex}(D_x, E_z) = D_x \rightarrow E_z \text{ with } \begin{cases} D_x = f(O_x) \\ E_z \subseteq E \end{cases} \quad (7d)$$

$$M_p(p,V) = G_p \rightarrow V \quad (7e)$$

In (7c), we map the set of qualifiers $Q_m$ to the set of entities $E$. According to (3) and (4a), the number of elements in $Q_m$ depends on $O_n$ so we designate set $Q_x$ as qualifiers and set $E_y$ as entities which remain after executing (7b). We will describe this matching in detail in Section 4.6.

In (7d), $O_x$ is an output set of objects that is successfully mapped from (7b). $D_x$ is a set of property values taken from WST-1 and WST-2 statements for each $o_i$ in $O_x$ by function $f(O_x)$. $E_z$ is a remaining set after executing (7b) and (7c). Section 4.7 will make clear about this matching.

Note that if every sub-step except the extra matching and the predicate matching fails, in consequences, the mapping will also fail. This is even applied for minor tasks in sub-steps.

*4.4. Subject matching*

In (7a), our task is to align between two sets $G_s$ and $U$ and then to find the matching pairs. The result can describe as the following formula:

$$M_s(s,U) = G_s \xrightarrow{1-many} U \quad (8)$$

$$= \{(g_i, u_j) \left| \begin{array}{l} \exists! g_i(tv_j = g_i), \\ tv_j \in u_j, \\ g_i \in G_s \end{array} \right. \}$$

A element $u_j$ in $U$ is mapped to a unique element $g_i$ in $G_s$ by literally matching its term value $tv_i$ (6b) to $g_i$. In contrast, $g_i$ can link to many elements $u_j$ in $U$.

In quad $T_i$, $s$ refers to the subject element or an item, and we can build mapping set $G_s$ from the formula (6a). In Wikidata, an item is classified into proper classes by setting values to its membership properties, such as *P31* (instance_of) or *P279* (subclass_of). If *P31* has value *Q5* (human), we need to add some values to $G_s$:
- *P735:given_name*: given name of *s*
- *P734:family_name:* family name of *s*

- *Q49614:nick_name*: nick name of *s*
- *P21:sex_or_gender:* We can add values, "*he*" or "*she*" depending the property value is "*male*" or "*female*".

In most of the world cultures, we refer a person by full name, last name, first name, nickname, gender (he or she), sometimes middle name and occupation title. With that reason, we must add some above values to ensure that having more chances to map successfully to the sentence subject. For example, in Figure 3 and Table 3, set $G_s$ for *Q1372810* is *{"Simone Loria", "Simone", "Loria", "he"}*. By using parsing to identify the sentence root, we are able to get sentence subjects and their positions, which usually appear before this root. In the data colletion process, we choose sentences having one subject so set U will contain only one element.

*4.5. Object matching*

Different from the subject matching, this matching is to match sequentially each element $o_i$ of set $O_n$ with set of entities $E$. Let start with mapping the first element $o_1$ in $O_n$ to $E$, here is the formula.

$$M_{o_1}(o_1, E) = G_{o_1} \xrightarrow{1-1} E \quad (9a)$$

$$= \{\exists!(g_i, e_i) \mid g_i = tv_i, g_i \in G_{o_1}, e_i \in E, tv_i \in e_i\}$$

$$= \{o_1, e_i\} = K_1 \text{ with } E_1 = \{e_i\}$$

We will find a unique pair *($g_i$, $e_i$)* that satisfy conditions in (9a). Because $g_i$ is an element of set $G_{o1}$ which is created from $o_1$, so we can replace it by $o_1$ to know which object that $g_i$ should map to. $K_1$ is a result set with only a matched pair *{($o_1$, $e_i$)}* and $E_1$ is a set containing only a matched entity. We define $K_1$ and $E_1$ as sets even though they contains only one element. The reason is there may have a case that we have more than one matched pair. If this happens, we will drop the whole mapping process to reduce the complication. The remaining entities of set $E$ is $E - E_1$. We continue with mapping second element $o_2$ to $E - E_1$ by the same formula.

$$M_{o_2}(o_2, E) = G_{o_2} \xrightarrow{1-1} E - E_1$$

$$= \{\exists!(g_j, e_j) \mid g_j = tv_j, g_j \in G_{o_2}, e_j \in E - E_1, tv_i \in e_i\}$$

$$= \{o_2, e_j\} = K_2 \text{ with } E_2 = \{e_j\} \quad (9b)$$

From (9a) and (9b), we can generalize the object matching between $O_n$ and $E$ as the formula.

$$M_O(O_n, E) = G_{O_n} \to E = (G_{o_1} \xrightarrow{1-1} E) \quad (9c)$$
$$\sqcup (G_{o_2} \xrightarrow{1-1} (E - E_1) \sqcup ...$$
$$\sqcup (G_{o_n} \xrightarrow{1-1} (E - E_1 - E_2 - ... - E_{n-1})$$
$$= \bigsqcup_{i=1}^{n} \left( G_{o_i} \xrightarrow{1-1} (E - \sum_{j=1}^{i-1} E_j) \right)$$

From (9c), we describe the result of the object matching as two formulas:

$$M_O(O_n, E) = K_1 \sqcup K_2 \sqcup ... \sqcup K_n \quad (9d)$$
$$= \bigsqcup_{i=1}^{n} K_i = K$$
$$K_i = G_{o_i} \xrightarrow{1-1} (E - \sum_{j=1}^{i-1} E_j) \quad (9e)$$

The remaining entities of $E$ is represented as $E_y$ in the formula:

$$E_y = E - E_1 - E_2 - ... - E_{n-1} = E - \sum_{i=1}^{n-1} E_i \quad (10a)$$
$$E_y = E - \{e_j \mid (o_i, e_j) \in K\} \quad (10b)$$

In a similar way, we have the matched objects $O_x$ with the formula:

$$O_x = \{o_i \mid (o_i, e_j) \in K\} \quad (11)$$

We will explain once again about the steps needed to do in the object matching. Starting with the first element $o_1$ in $O_n$, we look for the matched pair $K_1$ between of $G_{o1}$ and $E$ as well as the matched entity $E_1$. Then, we remove $E_1$ from $E$ or the remaining entity set now is $E$-$E_1$. This step is to guarantee that an element of entity set $E$ is only mapped to an element of object set $O$. We have no intention to handle a case when mutiple objects mapped to a single object or vice versa. After that, we do the same thing with second element $o_2$ and $E$-$E_1$ to find matched pairs $K_2$ and $E_2$. For element $o_n$, we will have the match pair $K_n$ and $E_{n-1}$. The result is a set of matched pairs *(o, e)* as $K$ shown in (9d) and (9e). To prepare for the qualifier matching, we define set $O_x$ with matched objects (11), which is used to create a set of qualifiers $Q_x$ from (3) and (4a); and set $E_y$ includes the remaining entities (10a, 10b).

*4.6. Qualifier matching*

From Section 4.5, we will receive two sets, $O_x$ and $E_y$. We define $Q_x$ from $O_x$ by (3) and (4a) as the formula.

$$Q_x = \{Y(o_1) \cup Y(o_2) \cup ... \cup Y(o_x) | \forall x \in N_1\} \quad (12)$$

The matching process is exactly akin to the subject matching in Section 4.5, which requires to match every qualifier $q_i$ in $Q_x$ to entity set $E_y$. From (9c), we have a similar general formula for the qualifier matching:

$$M_Q(Q_x, E_y) = G_{Q_x} \to E_y = (G_{q_1} \xrightarrow{1-1} E_y) \quad (13a)$$
$$\sqcup (G_{q_2} \xrightarrow{1-1} (E_y - E_1)) \sqcup ...$$
$$\sqcup (G_{q_x} \xrightarrow{1-1} (E_y - E_1 - E_2 - ... - E_{x-1}))$$
$$= \bigsqcup_{i=1}^{x} G_{q_i} \xrightarrow{1-1} (E_y - \sum_{j=1}^{i-1} E_j)$$

with $E_1, E_2, ..., E_{x-1}$ are matched entity sets for consecutive matching steps between $q_i$ and $E_y$.

We also define the result $L$, a set of matched pairs $(q, e)$ for this matching by the formula:

$$L = M_{Q_x}(Q_x, E_y) = L_1 \sqcup L_2 \sqcup ... \sqcup L_x \quad (13b)$$
$$= \bigsqcup_{i=1}^{x} L_i$$
$$L_i = G_{q_i} \xrightarrow{1-1} (E_y - \sum_{j=1}^{i-1} E_j) \quad (13c)$$

In (13c), $L_i$ are matched pairs between $q_i$ and $E_y$. We call the result for the qualifier matching is $L$ as (13b). Next, we denote $E_z$ as a set of remaining qualifiers:

$$E_z = E_y - E_1 - E_2 - ... - E_{x-1} = E - \sum_{i=1}^{x-1} E_i \quad (14a)$$
$$E_z = E_y - \{e_j | (q_i, e_j) \in L\} \quad (14b)$$

The qualifier matching depends on how many matched objects we obtain in the object matching. Normally in the experiment, there are just a few, one or two matched objects when matching from a single sentence. Each qualifier is also mapped to only one entity in $E_y$. After this matching, there are two sets, $O_x$ and $E_z$ should be prepared for the extra matching.

*4.7. Extra matching*

The extra matching is the process that maps remaining entities $E_z$ in a sentence to statements linked with elements (subjects, objects, and qualiliers) in a quad. In this paper, we focus on objects to extend more matchings. Except for deviring $E_z$ from the previous matching, we use dependency parsing to detect dependent phrases $DP$ (nouns or noun phrases) in a sentence. In Figure 4, object "*Bologna*" has phrases: "*Serie*", "*A*", and "*club*" as dependencies. We apply n-gram to produce more phrases such as, "*Serie A*", "*A club*", and "*Serie A club*" to gain a better matching. The longer phrase is matched is the more favorite.

We prefer to use $DP$ before $E_z$. For maximizing the matching capability, dependent phrases is detected by starting from sentence *root,* which must be a verb, e.g "*moved*" in Figure 3. Then, remove from $DP$ the matched phrases (subject, objects, qualifiers) to avoid conflicts. In Figure 3, we have $DP$ and $E_z$ are the same. However, in other cases, there will be a significant difference that helps to improve the matching performance.

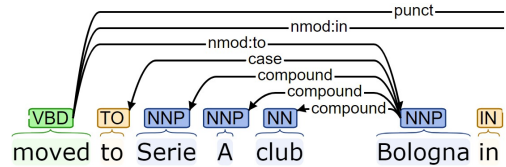

Fig. 4. Basic dependencies of object "club" in the example sentence in Figure 3, analyzed by Stanford CoreNLP[4].

From $O_x$ and $E_z$ in Section 4.6, we back to (7d) to define how to create set $D_x$ from $O_x$. For each $o_i$ in $O_x$, we use function $f(o_i)$ as a method to retrieve a set of WST-1 and WST-2 statements with respect to $o_i$, called $d_i$. Here is the definition of set $d_i$:

$$d_i = f(o_i) = \{(s_i, p, o) \left| \begin{array}{l} s_i = o_i, \\ o = pv \vee o = G_o \end{array} \right.\} \quad (15)$$

In (15), $o_i$ should be an Wikidata item and its WST-1 and WST-2 statement are represented in the form *($s_i$,*

---

[4] https://corenlp.run/

*p, o)* or *(subject, predicate, object)*. If tuple *(s_i, p, o)* is a WST-1 statement, we describe this tuple as *(s_i, p, pv)* with *pv* for the predicate value. Otherwise, if this tuple is a WST-2 statement, which the data type of *o* is wikibase-item. In that case, to map entities $E_z$ to *o*, we need to create $G_o$ with the formula in (6a).

From (15), we determine the formula for $D_x$ as:

$$D_x = f(O_x) = f(o_1) \sqcup f(o_2) \sqcup ... \sqcup f(o_x) \quad (16)$$
$$= d_1 \sqcup d_2 \sqcup ... \sqcup d_x$$

Analogous to the object matching and the qualifier matching, we also map each element of $D_x$ to $E_z$, we would like to present the extra matching (7d) in a general formula as:

$$M_{Ex}(D_x, E_z) = D_x \rightarrow E_z = (d_1 \rightarrow E_z) \quad (17a)$$
$$\sqcup (d_2 \rightarrow (E_z - E_1)) \sqcup ...$$
$$\sqcup (d_x \rightarrow (E_z - E_1 - E_2 - ... - E_{x-1}))$$
$$= \bigsqcup_{i=1}^{x} d_i \rightarrow (E_z - \sum_{j=1}^{i-1} E_j)$$

$$J = M_{Ex}(D_x, E_z) = J_1 \sqcup J_2 \sqcup ... \sqcup J_x = \bigsqcup_{i=1}^{x} J_i \quad (17b)$$

$$J_i = d_i \rightarrow (E_y - \sum_{j=1}^{i-1} E_j) \quad (17c)$$

with $E_1, E_2, ..., E_{x-1}$ are matched entity sets for consecutive matching steps between $d_i$ and $E_z$. We denote *J* (17b) is the total result and $J_1, J_2, ..., J_n$ are component results for each matching step.

Next, we expand the formula (17c) to clarify matched pairs needed to collect as:

$$J_i = d_i \rightarrow (E_y - \sum_{j=1}^{i-1} E_j) \quad (17d)$$
$$= \left\{ (s_i, p, o, e_i) \middle| \begin{array}{l} \exists! e_i \therefore tv_i = pv, tv_i \in e_i, \\ pv \in G_o \vee pv = o, \\ e_i \in \{E_y - \sum_{j=1}^{i-1} E_j\} \end{array} \right\}$$

Once again, to avoid the conflicts, we only map an entity to a unique tuple *(s_i, p, o)*. Back to Figure 3, we replace $e_i$ by *p*, not by *o* in creating the labelled sentence (*P118* and *P31*) because *o* can take any values. We care more about element *p* to see how far extra matchings can go.

### 4.8. Predicate matching

To continue with the predicate matching, we need to receive in hand two sets, $G_p$ formed by (6a) and *V* retrieved from the syntactic analysis. As mentioned in Section 4.1 and Section 4.3, there is a comparison gap between these sets, where $G_p$ contains phrases (mainly nouns or noun phrases) and *V* is a set of verbs. This leads to a very small chance that $G_p$ and *V* has something in common.

$G_p$ is a set of label and aliases for element *p* as a predicate in quad $T_i$ *(s, p, $O_n$, $Q_m$)*. Expressing theoretically a relation of subject *s* and objects *o* in RDF triples *(s, p, o)*, our intuition points that element *p* may describe the relation between subject *s* and objects $O_n$, and even $Q_m$. However, we will discuss this discovery later in Section 6.4.

To assure the matching ability, we widen $G_p$ to $B_p$, which adds verbs to compare to set *V*. So we now have the general formula:

$$M_p(p, V) = G_p \rightarrow V = B_p \rightarrow V \quad (18)$$

Our attention turns to how to form $B_p$ to adapt the requirement. We introduce a minor step to rebuild $G_p$ in a fuction *sw* which consists of a set of words (except stop words) called $C_p$ without any processing tasks like lemmatization or stemming.

$$C_p = sw(G_p) = \{c_1, c_2, ..., c_n \mid n \in N_1\} \quad (19)$$

For each word $c_i$ in *C*, we look up top *t* nearest words to it by vector distance in a pre-trained word vector model, such as *wiki-news-300d-1M*[5] or *GoogleNews-vectors-negative300.bin*[6]. We define set *Z* for top *t* words near $c_i$ as:

$$Z(c_i, t) = \{(w_1, d_1), (w_2, d_2), ..., (w_t, d_t)\} \quad (20)$$

Next, we call $B_p$ is the union set of all sets $Z(c_i, t)$ corresponding every element in *C*.

$$B_p = Z(c_1, t) \cap Z(c_2, t) \cap ... \cap Z(c_n, t) \quad (21)$$
$$= \bigcup_{i=1}^{n} Z(c_i, t)$$
$$= \{(w_i, d_{max}) \mid d_{max} = \max(d_1, d_2, ... d_z)\}$$

---

[5] https://fasttext.cc/docs/en/english-vectors.html
[6] https://code.google.com/archive/p/word2vec/

In (21), $B_p$ consists of $(w_i, d_{max})$ pairs, whose $d_{max}$ is the highest vector distance in the scale (0-1) of those having the same $w_i$. For example, a group of pairs: *("wedding", 0.77), ("wedding", 0.56), ("married", 0.65)* will become: *("wedding", 0.77), ("married", 0.65)* through out the formula (21). At last, we will extend (18) as the final matching formula:

$$M_p(p,V) = G_p \to V = B_p \xrightarrow{1-1} V \quad (22)$$

$$= \left\{ (w_i, d_{max}, e_j) \middle| \begin{array}{l} tv_j = w_i, \\ tv_j \in e_j, e_j \in V, \\ (w_i, d_{max}) \in B_p \end{array} \right\}$$

In (22), each element (verb) in $V$ is only needed to map to an element in $B_p$ by a distance $d$ to prove it in somehow related to the element $p$ and quad $T_i$. Set $B_p$ is not always expanded from $G_p$. We can extend it from the combination of $G_s$, $G_O$ (object instances), and $G_Q$ (qualifier instances) depending on cases to receive better results.

By our view, this matching process is a problem of classification or clustering. Since then, we do not apply the predicate matching in the mapping process because we would like to discover how many verbs possible are there for a given Wikidata statement. Since then, we are able to learn relationships between predicate $p$ and other elements in quads $T_i$. Verbs will be applied some features (IDF, TF, and vector distances) and may cluster in groups to check whether they fit a given quad or not.

## 5. Data Collection

We utilize the interwiki link structure built to connect Wikidata and other Wikimedia projects, including English Wikipedia in the data collection process. To gain the uniformity, English is used for both Wikipedia page content and Wikidata statement's labels. However, this paper can apply to any language if possible because Wikipedia and Wikidata are multilingual projects.

### 5.1. Data collection

Figure 5 simulates a sketch model, which is used for the mapping process. From a page in English Wikipedia to an item in Wikidata, there may not exist any interwiki link or exists only one. Suppose that the latter case happens, we record this pair in the form *(item, page)*. Next, a minor code will deserialize the XML or JSON data of this item to extract its statements, then encode them as a set of quads and triples as mentioned in Section 3.5. From that, we continue to detach quads $T_n$ in the form $(s, p, O_n, Q_m)$. In the other side, we process the page content, formatting and organizing it to a set of sentences $S$.

Now, we define the mapping process for item $it_i$ and a page $p_j$ as the formula:

$$M(it_i, p_j) = M(T_n, S) = T_1 \to S \quad (23)$$
$$\sqcup T_2 \to S - S_1 \sqcup ... \sqcup$$
$$\sqcup \cup T_n \to S - S_1 - S_2 - ... - S_{n-1}$$
with $n \in N_1, S \neq \emptyset$

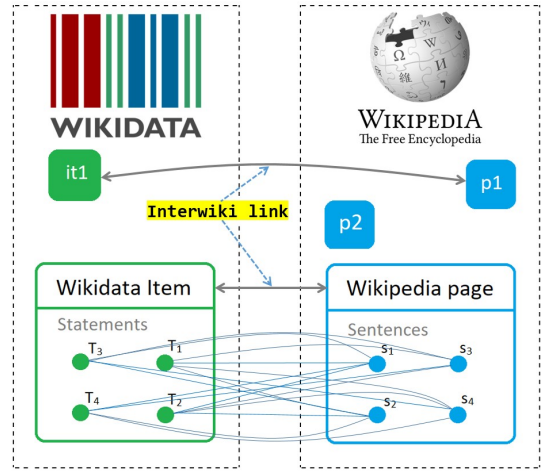

Fig. 5. The sketch how to collect and map data between Wikidata and Wikipedia. Every quad $T_i$ of a Wikidata item is mapped to every sentence $s_j$ of a Wikipedia page.

In (23), this process is chopped to smaller mapping steps for every $T_i$ and $S$, which is specified in (5). Set $T_n$ and set S must be not empty. If these conditions are not satisfied, we will discard the mapping. $S_1$, $S_2,...,S_n$ are results for each mapping step for $T_1, T_2,...,T_n$ and remaining sets $S, S-S_1,..., S-S_1-S_2-...S_{n-1}$. This allows avoiding that the multiple sentences mapped to a quad $T_i$ although it is practically rare to occur, except when the mapping is executed with the first paragraph of pages, where may occur a high semantic density. At last, we form a formula for mapping a set of *(item, page)* pairs as:

**Query 1:** Retrieve all qualifiers of property *P26* and count their occurrence frequency.

```
 1:  SELECT ?qual ?qualLabel ?count WHERE {
 2:    {
 3:      SELECT ?qual (COUNT(DISTINCT ?item) AS ?count) WHERE {
 4:        hint:Query hint:optimizer "None" .
 5:        ?item p:P26 ?statement .
 6:        ?statement ?pq_qual ?pq_obj .
 7:        ?qual wikibase:qualifier ?pq_qual .
 8:      }
 9:        GROUP BY ?qual
10:    } .
11:
12:    OPTIONAL {
13:      ?qual rdfs:label ?qualLabel filter (lang(?qualLabel) = "en") .
14:    }
15:  }
16:  ORDER BY DESC(?count) ASC(?qualLabel)
```

**Query 2:** Capture a page list by qualifiers (*P580*, *P582*) of property *P26*.

```
 1:  SELECT ?item ?title ?object ?property ?value ?sitelink WHERE {
 2:      ?item p:P26 ?object.
 3:      ?object ps:P26 ?property;
 4:              pq:P580|pq:P582 ?value.
 5:      ?sitelink schema:about ?item;
 6:        schema:isPartOf <https://en.wikipedia.org/>;
 7:        schema:name ?title.
 8:      SERVICE wikibase:label { bd:serviceParam wikibase:language "en,en". }
 9:    }
10:  LIMIT 50000
```

Table 7

List of properties used for the experiments

| Property | Label | Description | Aliases | Size of page list |
|---|---|---|---|---|
| P26 | spouse | the subject has the object as their spouse (husband, wife, partner, etc.), use "partner" (P451) for non-married companions | wife, married to, consort, partner, marry, marriage partner, married, wedded to, wed, life partner, wives, husbands, spouses, partners, husband | 23294 |
| P39 | position held | subject currently or formerly holds the object position or public office | political office held, political seat, public office, office held, position occupied, holds position | 22246 |
| P54 | member of sports team | sports teams or clubs that the subject currently represents or formerly represented | member of team, team, team played for, played for, plays for, teams played for, of team, club played for, player of, part of team | 7906 |
| P69 | educated at | educational institution attended by subject | alma mater, education, alumni of, alumnus of, alumna of, college attended, university attended, school attended, studied at, graduate of, graduated from, faculty | 19214 |
| P108 | employer | person or organization for which the subject works or worked | workplace, employed by, works at, working for, worked for, works for, worked at, working place | 10899 |
| P166 | award received | award or recognition received by a person, organisation or creative work | prize received, awards, honorary title, recognition title, award, honours, honors, medals, awarded, award won, prize awarded, awards received, win, winner of | 30354 |

$$M(it_n, p_n) = M(it_1, p_1) \sqcup M(it_2, p_2) \sqcup \quad (24)$$
$$\sqcup ... \sqcup M(it_n, p_n), n \in N_1$$

Formula (24) indicates the most general method for the mapping process of *n (item, page)* pairs in this research. In the next section, we will present a quick way to collect properly these pairs via SPARQL queries at Wikidata server.

*5.2. Item-page pair selection*

In *List of Properties/all*[7], we look up some properties that satisfy two criteria: to obtain a high number of uses and to have values (objects) with "*WikibaseItem*" data type. The latter helps a property more opportunities to connect it to its qualifiers and qualifier values that allow shaping quads *(s, p, $O_n$, $Q_m$)* in the mapping process. For every chosen property, a SPARQL query is responsible for checking the linked qualifiers and also their number of uses. Query 1 is used to find qualifiers of property *P26* while Table 8 shows the result of this query.

Table 8

Some qualifiers of property *P26*.

| Qualifier | Qualifier label | Number of uses |
|---|---|---|
| wd:P580 | start time | 32543 |
| wd:P582 | end time | 20093 |
| wd:P2842 | place of marriage | 2004 |
| wd:P1534 | end cause | 1456 |
| wd:P1545 | series ordinal | 534 |
| wd:P585 | point in time | 239 |
| wd:P1971 | number of children | 198 |
| … | … | … |

Based on the number of uses, we continue to select some qualifiers as parameters in the next SPARQL query to extract *(item, page)* pairs. In Table 8, two qualifiers, *P580* and *P582* are the most dominant qualifiers of property *P26* so we take them to be parameters for Query 2. This query will offer a set of tuples in the form *(item, lemma, object, property, value, sitelink)*, which continues to be extracted to get *(item, lemma)* or *(item, page)* pairs.

Wikidata Query Service[8] provides a web interface to run SPARQL queries such as *Query 1* and *Query 2*. After running manually these two queries and check out their outputs, we pick out some experimental properties, such as *P26*, *P39*, *P54*, *P69*, *P108*, and *P166* with detailed information is shown in Table 7. In total, there are 133913 of *(item, page)* pairs that we have to proceed in the scanning process. Finally, we store these properties in corresponding *.csv* files by their names as inputs for the data scanning.

*5.3. Data scanning*

Instead of massive database dumps, English Wikipedia API[9] and Wikidata API[10] are used to access instantly online entities (items/pages). For each property in Section 5.2, our code will read its *.csv* file to yield a list of *(item identifier, page name)* pairs. Later, URL queries take item identifier and page name values in every pair as parameters to call requests to APIs and wait for data return in the XML/JSON format. In each pair, we will receive from its page name and identifier, in turn, are HTML content and statement structure.

To handle the page content is uncomplicated with several steps of text preprocessing: remove HTML markups, special characters, trim redundant spaces, and split into sentences. However, it is a bit sophisticated to convert to list of statements *T* from the statement structure. We would like to present Algorithm 1, which is about how to extract statements from the return XML data of an item.

In Algorithm 1, from an input *id*, we find its XML *root* in Wikidata. Then, loop each exists *predicate* in *root* to get predicate (property) identifier *p* and continue to loop each *claim* by *predicate* to get predicate data type *pt*. Once again, loop each *object* in *claim.objects* to get predicate (object) value *pv*. If *pt* equals to *"wikibase-item"*, it means predicate value *pv* is a Wikidata item, so we can set object *ob* equal to *pv*. We check the qualifier set *qualifiers* of *pv* or *ob* to decide what to do in the next step. If the number of *qualifiers* is zero or this set is empty, the statement will be either WST-1 or WST-2. If *ob* is empty (*pt* takes value *"wikibase-item"*), the statement will be WST-1 as *r1(s, p, pt, pv)*, otherwise WST-2 as *r2(s, p, pt, ob)*, before adding *('r1', r1)* and *('r2', r2)* to *claim_list*. We add *pt* to these tuples to recognize the data type of predicate without separating to a new triple like *(p, :data_type, pt)*. If *qualifiers* is not empty, we temporary set that this statement is WST-3. Next, loop each *qualifier* in *object* to get qualifier *q* and qualifier data type *qt*. We deal with multiple

---

[7] https://www.wikidata.org/wiki/Wikidata:Database_reports/
[8] https://query.wikidata.org/
[9] https://en.wikipedia.org/w/api.php
[10] https://www.wikidata.org/w/api.php

**Algorithm 1:** Extract statements of an item in the XML data format from Wikidata API.

**Input:** item identifier: *id*
**Output:** list of statements: *claim_list*

```
1:  root := get wikidata root by id, claim_list := empty
2:  s := id //subject (item identitifer)
3:  p, ob, pt, pv, q, qt, qv := empty
4:  loop predicate in root
5:      p := get predicate (property) identifier
6:      loop claim in predicate
7:          pt := get predicate data type
8:          pv := get predicate value in claim.objects
9:          if (pt is 'wikibase-item')
10:             ob := pv
11:         end if
12:         qualifiers := get qualifiers
13:         if (qualifiers is empty)
14:             if (ob is empty):
15:                 r1 := (s, p, pt, pv) // WST-1
16:                 add ('r1', r1) to claim_list
17:             else
18:                 r2 := (s, p, pt, ob) // WST-2
19:                 add ('r2', r2) to claim_list
20:             end if
21:         else
22:             if (pv is not empty)
23:                 r3 := (s, p, pt, pv)
24:             else
25:                 r3 := (s, p, pt, ob)
26:             end if
27:             loop qualifier in qualifiers
28:                 q := get qualifier identifier
29:                 qt := get qualifier data type
30:                 q_item := (q, qt)
31:                 loop value in qualifier:
32:                     qv := get qualifier value
33:                     if (qv is not empty):
34:                         add qv to q_item
35:                         add q_item to r3
36:                     end if
37:                     reset qv to empty
38:                 end loop
39:                 reset q, qt, q_item to empty
40:             end loop
41:             if (len(r2) > 4)
42:                 add ('r3', r3) to claim_list // WST-3
43:             else
44:                 if (pt is not 'wikibase-item')
45:                     add('r1',r3) to claim_list // WST-1
46:                 else
47:                     add ('r2',r3) to claim_list // WST-2
48:                 end if
49:             reset ob, pv to empty
50:         end loop
51:         reset pt to empty
52:     end loop
53:     reset p to empty
54: end loop
55: return claim_list
56:
57:
```

qualifiers by looping each *value* in *qualifier* to get a set of qualifier values *q_item*, lastly add to *r3*. If *r3* contains qualifiers (whenever its length is larger than 4), it is a WST-3 statement. If not, we check the *pt* value. If *pt* is not '*wikibase-item*', this statement is WST-1, otherwise WST-2.

For example, in Figure 3, we extract the WST-3 statement that covers quad *('Q1372810', P54, Q1893, P580)* in the form:

('r3',
 ('Q1372810', 'P54', 'Q1893', 'wikibase-item',
   ('P580', 'time', '+2011-01-01T00:00:00Z'),
   ('P582', 'time', '+2012-01-01T00:00:00Z'),
   ('P1350', 'quantity', '+9'),
   ('P1351', 'quantity', '+1'),
   ('P1642', 'wikibase-item', 'Q3622633')
 )
)

The above statement is in the form *(s, p, o, $Q_m$)*, if we combine many WST-3 statements together, we will have the form *(s, p, $O_n$, $Q_m$)*, exactly like the formula (1).

We apply Algorithm 2 to scan *(item_identifier, page_name)* pairs classified by properties one by one. From a property identifier *pi*, we take its *page_list*, extracted from the file *{pi}.csv* to retrieve *(item_identifier, page_name)* pairs. The next step is to loop each pair in *page_list* to get *sentence_list* and *claim_list* (Algorithm 1) before executing the mapping between them (Algorithm 3) to receive *mapped_results*. Then, we save *mapped_results* to file *output_{pi}.csv* if it is not empty.

For the mapping between *sentence_list* and *claim_list* in Algorithm 2, we describe in detail in Algorithm 3. Loop each *sen* in *sentence_list* to extract three sets, *S* (subjects), *V* (verbs) and *E* (entities). Next, we receive $M_s$ as a matching subject pair between *S* and *claim_list*. Because we work with sentences with a single subject so $M_s$ will hold either one pair $(g_i, u_j)$ or empty. If $M_s$ is empty, the mapping is discarded and continue the loop with a new sentence. Otherwise, we will continue with the object matching. $M_o$ is a set of object maching pairs between *E* and *claim_list*. We also drop this mapping in case that $\overline{M_o}$ is empty. Otherwise, when it is successful, we will receive $O_x$ and $E_y$ as matched objects and remaining entities that we already removed the matched entities from *E*. Before working with the qualifier matching, we create set $Q_x$ by getting quali-fiers from $O_x$. Then, $M_p$ is a set of qualifier matching pairs and $E_z$ is a set of remaining entities. Likewise $M_s$ and $M_o$, if $M_p$ is empty, we also drop the matching and repeat a new loop.

---

**Algorithm 2:** Scan all *(item_identifier, page_name)* pairs by property names, search for matching results, and write these results to an output *.csv* file.

**Input:** property identifier: *pi*
**Output:** output file: *output_{pi}.csv*

1:   page_list := *get page list from file {pi}.csv*
2:   **loop** item_identifier, page_name **in** page_list
3:     sentence_list := *get sentences by page_name*
4:
5:     // Algorithm 1
6:     claim_list := *get claims by item_identifer*
7:
8:     // Algorithm 3
9:     mapped_results := *map btw claim_list & sentence_list*
10:
11:    **if** (mapped_results **is not** empty):
12:       *save mapped_results to file output_{pi}.csv*
13:    **end if**
14:    *empty mapped_results*
15:  **end loop**

---

Next is for the extra matching. Firstly, we check the number of elements in $E_z$, if it is empty, we will extend by collecting more items from the dependency parsing. From $O_x$ (a set of matched objects), we get all WST-1 and WST-2 statements of each element in $O_x$ and combine them, called $D_x$. $M_e$ is the mapping results between $D_x$ and $E_z$. Then, we create labeled sentence *ls* and add tuple *(sen, ls, $M_s$, $M_o$, $M_p$, $M_e$)* to the result. Finally, we receive 18510 output sentences over 113913 scanned pages or the rate of 0.16 sentences per page which shows that it is a very rare chance to get the required sentences.

The data scanning process may take a long time, from several days up to a week because it works through online APIs of Wikipedia and Wikidata. For that reason, it is wiser to save mapped results to files in every scanning step to avoid some unpredictable incidents that may happen in the going. We assume to reduce the executed time up to less than 24 hours if we use database dumps downloaded from the servers. Unfortunately, at the moment of data collection, we lack computer power to deal efficiently with reading hundreds of gigabyte data of Wikidata and Wikipedia dumps.

---

**Algorithm 3:** Map sentences to statements (claims) to find mapped results.

**Input:** a list of sentences: *sentence_list*, a list of statements: *claim_list*
**Output:** results: *mapped_results*

1:   mapped_results := [] // set to empty
2:   **loop** sen **in** sentence_list
3:     // S: subjects, V: verbs, E: entities
4:     S, V, E := *get subjects, verbs, entities by sen*
5:
6:     // Ms: subject matching pairs
7:     Ms := *match subject between S and claim_list*
8:     **if** (Ms **is** empty)
9:        *drop mapping & continue new loop*
10:    **end if**
11:
12:    // Mo: object matching pairs
13:    // Ox: matched objects, Ey: remaining entities
14:    Mo, Ox, Ey := *match objects btw E and claim_list*
15:    **if** (Mo **is** empty)
16:       *drop mapping & continue new loop*
17:    **end if**
18:
19:    // Mp: qualifier matching pairs
20:    // Ez: remaining entities
21:    Qx := *get qualifiers by Ox*
22:    Mp, Ez := *match qualifiers between Qx and Ey*
23:    **if** (Mp **is** empty)
24:       *drop mapping & continue new loop*
25:    **end if**
26:
27:    **if** (Ez **is** empty)
28:       Ez := *get dependency nouns/noun phrases*
29:    **end if**
30:
31:    // Me: extra matching pairs (optional)
32:    // Dx: extra statements of Ox
33:    Dx := *get WST-1 and WST-2 statements of Ox*
34:    Me := *match statements between Dx and Ez*
35:
36:    ls := *create a labeled sentence by result pairs*
37:    *add (sen, ls, Ms, Mo, Mp, Me) to mapped_results*
38:    **end if**
39:  **end loop**
40:  **return** mapped_results

---

## 6. Corpus evaluation

### 6.1. Mapping result evaluation

In this section, we evaluate output sentences and their NeuralCoref[11] texts after training on the whole pages where they belongs to. We use five entity link-ing methods of data matching (AIDA, OpenTapioca, TAGME, WAT, and Wikifier) and two methods of type matching of NERs (DBpedia and Wikidata) to evaluate sentences in three independent components: subject matching, object matching and qualifier matching. For each sentence, we calculate the matching scores applied both for type matching and data matching.

---

[11] https://spacy.io/universe/project/neuralcoref/

We evaluate two matching scores: $sc_1$ is a score of successful matchings over the total number of matchings, and $sc_2$ is the score of successful matchings over the successful mappings to Wikidata/DBPedia. We form a simple formula,

$$sc_1 = \frac{N_s}{N}, \quad sc_2 = \frac{N_s}{N - N_u} \quad (25a, 25b)$$

where $N$ is the total number of matchings while $N_s$ is the number of successful matchings, and also successful mappings to Wikidata/DBpedia. $N_u$ is the number of mappings that unsuccessful link to Wikidata/DBpedia, in a result as unsuccessful matchings. $N_u$ represents the drawback of entity linking algorithms rather than our evaluation method.

Table 9 shows results of the type matching of sentence components. We use the hierachy structure of two Wikidata properties: *P31 (instance of)* and *P279 (subclass of)* to form the hypernym tree. The hypernym level is denoted *lvx* where *x* is the tree level compared to the current node. For example, *lv0* is no need to get hypernyms while *lv1* is to get the direct hypernyms of the current node. The higher level is presumed to gain more chance to match data types successfully.

Table 9

Type matchings of NERs on subjects, objects and qualifiers.

|  | Subject matching | Object matching | Qualifier matching |
| --- | --- | --- | --- |
| **wd-sc1-lv0** | 0.5848 | 0.2100 | 0.9355 |
| **wd-sc2-lv0** | 0.5848 | 0.2100 | 0.9355 |
| **wd-sc1-lv1** | 0.5848 | 0.2456 | 0.9355 |
| **wd-sc2-lv1** | 0.5848 | 0.2457 | 0.9355 |
| **wd-sc1-lv2** | *0.5848* | 0.3549 | 0.9355 |
| **wd-sc2-lv2** | *0.5848* | *0.3551* | 0.9355 |
| **db-sc1-lv0** | 0.2662 | 0.4092 | - |
| **db-sc2-lv0** | 0.5892 | 0.6331 | - |
| **db-sc1-lv1** | 0.2586 | 0.4092 | - |
| **db-sc2-lv1** | 0.5864 | 0.6332 | - |
| **db-sc1-lv2** | 0.2662 | *0.4096* | - |
| **db-sc2-lv2** | 0.5892 | 0.6333 | - |
| *wd: Wikidata, db: DBpedia, sc: score, lv: level* | | | |

Compare to Wikidata, the scores mapped to DBpedia are generally higher because this project extracts automatically factual information from Wikipedia and also has many methods to verify the result data when Wikidata works mostly as a human collaboration model. For subject matching and object matching, the best performance belongs to a DBpedia method (*db-sc2-lv2* with the scores respectively as *0.5892, 0.6333*), however, these scores of Wikidata are stable in the case of subject and increase steadily by levels in the case of object matching. For qualifier matching, DBpedia lacks the support of qualifier data types so Wikidata is the only method. This matching obtains a score of *0.9355* when the qualifiers include several basic data types (datetime, number, etc) so likely easier to gain a high score. Furthermore, we can see that the levels of hypernyms have no effect on the matching performance.

Table 10

Data matchings of NERs on subjects and objects.

|  | Subject matching | Object matching |
| --- | --- | --- |
| **utr-aida-sc1** | 0.1995 | 0.3713 |
| **utr-aida-sc2** | 0.2154 | 0.4008 |
| **utr-opentapioca-sc1** | 0.0708 | 0.4761 |
| **utr-opentapioca-sc2** | 0.0708 | 0.4762 |
| **utr-tagme-sc1** | *0.2266* | *0.7596* |
| **utr-tagme-sc2** | *0.2266* | *0.7597* |
| **utr-wat-sc1** | 0.2036 | 0.6733 |
| **utr-wat-sc2** | 0.2076 | 0.6865 |
| **utr-wikifier-sc1** | 0.1864 | 0.5011 |
| **utr-wikifier-sc2** | 0.1864 | 0.5011 |
| **tr-aida-sc1** | 0.2828 | - |
| **tr-aida-sc2** | 0.2964 | - |
| **tr-opentapioca-sc1** | 0.0440 | - |
| **tr-opentapioca-sc2** | 0.0440 | - |
| **tr-tagme-sc1** | 0.3250 | - |
| **tr-tagme-sc2** | 0.3250 | - |
| **tr-wat-sc1** | 0.3267 | - |
| **tr-wat-sc2** | 0.3291 | - |
| **tr-wikifier-sc1** | **0.3373** | - |
| **tr-wikifier-sc2** | **0.3373** | - |
| *tr: trained, utr: untrained* | | |

Next, we measure *sc1* and *sc2* again but now for the data values of subjects and objects. We apply two type of texts (trained and untrained) with five entity linking methods which mentioned above. In experiments, these methods usually some limits on detecting qualifiers so we do not use qualifier matching. According to Section 4.4 and Section 4.5, the subject matching also contains cases as human pronouns (he, she). Thus, that is reasonable when using NeuralCoref to get the trained texts to boost the matching per-formance. In contrast, since we have never applied any human pronoun in the object matching so there will be no trained texts here.

In Table 10, TAGME gains the best performance for untrained texts on subject matching and object matching while Wikifier is the winner for trained texts on subject matching. There are two explanations why we do have the low scores on subject matching. First, the reason may be from the limitation of neural networks using in the NeuralCoref pipeline as well as entity linking methods when they may impossible detect correct results in some cases. Second, Wikidata may offer incomplete, incorrect or missing structured data in the matching process because its data has been built from the willingness of contributors.

Refer to Table 9 and Table 10, the concurrence probabilities calculated by multiplying of the highest score (probability) of sentence components have the results as *0.3490* and *0.2562* respectively for type matching and subject matching. It roughly means one third or one quarter of the outcome sentences may guarantee the matching precision and this is usually accepted for the unsupervised problems. These results even are higher than our expectation when the language diversity is the main barrier of Data2Text issues which pulls down sharply the total performance. In that sense, the chance of mapping a required sentence in a bunch of random texts is relatively low.

*6.2. Some basic statistics*

This section lists some of the statistical methods on datasets. *Commons* is a dataset that includes six child datasets: *P26s*, *P39s*, *P54s*, *P69s*, *P108s*, and *P166s*. *Candidates* contain only candidate sentences (best sentences) extracted from *Commons*. We define a candidate sentence by criteria: (+) the number of redundant words of its labeled sentence must be only one; (++) this redundant word must be the sentence's root. In detail, (++) is the predicate (main verb) that we mentioned about the predicate matching in Section 4.8. *Candidates* is presumed as the nearly-gold corpus before applying the predicate matching and other evaluation methods to produce the result sentences.

To get the redundant words, we parse the labeled sentence by tokens and then remove tokens that appear in the stopword list (available in spaCy and NLTK packages) and our own stopword list. The latter contains tokens whose *token.pos_* are *X*, *PUNCT*, *CCONJ*, *ADP*, *PRON*, *PART*, *DET* as well as *token.dep_* is *punct*.

For example, we have this sentence from *P26s:*

Raw sentence:
  She and David Birney divorced in 1989.
Labeled sentence (also the candidate sentence):
  [s] and [o0] divorced in [o0:P582-qualifier] .

In this example, the labeled sentence has a redundant word is "*divorced*" and it is also the root. The other words, such as "*and*" and "*in*" are not be counted because they are stopwords.

Table 11 displays basic statistics on datasets which tabulate into 5 features:
  – *feature1*: The average of sentence length.
  – *feature2*: The average of the number of words per sentence.
  – *feature3*: The average of the number of tokens per sentence.
  – *feature4*: The ratio of token per quad.
  – *feature5*: The ratio of token per quad item.

The highest scores on all features take *Candidates* as the champion because it is designed with the purpose of choosing the best sentences without any redundant words in the mapping results. Its *feature1* and *feature2* values are *50.57* and *8.72* poses that generally, the outcome sentences are not too long. Moreover, *feature5* (the number of tokens between two squad items) is only *1.1* compared to *7.27* of *Commons* let us know that we must prune about 6.17 words per quad item to receive the candidate sentences. As our prediction, the volume of *Candidates* is also the least, with *1.59%* meanwhile the highest volume is *P54s* with *38.38%*. Except for *Commons* and *Candidates*, the best performance on all features is *P26s* and *P166s* is the worst.

*6.3. Noise filtering*

With each labeled sentence in the output, we extract its redundant words as set $N=\{w_1, w_2, ..., w_N\}$ and its quad items (Table 6) as set $Q=\{q_1, q_2, ..., q_Q\}$ to set up some distances (similarities) between them, based on values such as TF, IDF, local/global Word2vec, global Word2vec with qualifier, and their combinations. We then apply some density-based clustering algorithms (DBSCAN, OPTICs, LOF, etc) to filter outliers or noises out of the corpus. We deploy two Word2vec models, the local model trained

Table 11

Some basic statistics on different datasets.

|  | Commons | Candidates | P26s | P39s | P54s | P69s | P108s | P166s |
|---|---|---|---|---|---|---|---|---|
| Avg. Sentence length | 111.29 | **51.52** | 90.24 | 126.96 | 107.04 | 116.2 | 116.78 | 134.7 |
| Avg. Word per Sentence | 18.97 | **8.8** | 14.87 | 21.9 | 19.03 | 18.96 | 19.24 | 22.4 |
| Avg. Token per Sentence | 21.99 | **9.41** | 18.06 | 25.08 | 21.91 | 21.61 | 21.98 | 25.71 |
| Ratio of Token per Quad | 36.21 | **8.03** | 29.69 | 40.20 | 36.56 | 35.81 | 36.48 | 40.99 |
| Ratio of Token per Quad item | 7.27 | **1.11** | *6.06* | 7.9 | 7.25 | 7.3 | 7.42 | *8.29* |
| Volume | 100% | **1.59%** | 21.12% | 8.91% | **38.38%** | 11.19% | 8.24% | 12.12% |

Table 12

Evaluation criteria by four methods (cm1, cm2, cm3, cm4) and seven clustering algorithms.

| Methods | Executed time (s) | Noise rate (%) | F-M | ACC | Purity | Silhouette |
|---|---|---|---|---|---|---|
| cm1-nearestneighbors02 | 0.2114 | 0.09 | 0.9322 | **1.0** | 0.9990 | 0.9209 |
| cm1-dbscan01 | 3.8546 | 0.23 | 0.9307 | 0.9283 | 0.9976 | 0.7879 |
| cm1-optics01 | **25.8753** | 0.49 | 0.9280 | **1.0** | 0.9950 | 0.8368 |
| cm1-agglomerative | 17.6788 | 0.03 | 0.9328 | 0.9303 | 0.9996 | 0.9479 |
| cm1-localoutlier04 | 8.1384 | 0.05 | 0.9326 | 0.9306 | 0.9994 | 0.9394 |
| cm1-birch01 | 1.3643 | 0.03 | 0.9328 | 0.9303 | 0.9996 | 0.9479 |
| cm1-kmeans | 0.2114 | 27.86 | 0.7033 | 0.6520 | 0.7213 | **0.5728** |
| cm2-nearestneighbors02 | **0.1057** | 0.04 | 0.9328 | **1.0** | 0.9995 | 0.9453 |
| cm2-dbscan01 | 2.0405 | 0.11 | 0.9319 | **1.0** | 0.9988 | 0.9152 |
| cm2-optics01 | 21.5236 | 0.32 | 0.9298 | **1.0** | 0.9967 | 0.8647 |
| cm2-agglomerative | 12.1704 | 0.03 | 0.9328 | 0.9303 | 0.9996 | 0.9483 |
| cm2-localoutlier04 | 6.6906 | 0.05 | 0.9326 | 0.9306 | 0.9994 | 0.9400 |
| cm2-birch01 | 0.7393 | 0.03 | 0.9328 | 0.9303 | 0.9996 | 0.9483 |
| cm2-kmeans | 0.1964 | 27.84 | 0.7034 | 0.6521 | 0.7215 | 0.5807 |
| cm3-nearestneighbors02 | 0.1237 | 0.04 | 0.9328 | **1.0** | 0.9995 | 0.9368 |
| cm3-dbscan01 | 2.5701 | 0.08 | 0.9324 | **1.0** | 0.9991 | 0.9181 |
| cm3-optics01 | 22.9761 | 0.39 | 0.9290 | **1.0** | 0.9960 | 0.8346 |
| cm3-agglomerative | 13.7007 | 0.03 | 0.9329 | 0.9303 | 0.9996 | 0.9446 |
| cm3-localoutlier04 | 6.6310 | 0.05 | 0.9326 | 0.9306 | 0.9994 | 0.9320 |
| cm3-birch01 | 0.6918 | 0.08 | 0.9323 | 0.9298 | 0.9991 | 0.9173 |
| cm3-kmeans | 0.1874 | **29.25** | **0.6964** | 0.6380 | 0.7074 | 0.5824 |
| cm4-nearestneighbors02 | **0.1057** | 0.04 | 0.9327 | **1.0** | 0.9995 | 0.9457 |
| cm4-dbscan01 | 2.5611 | 0.10 | 0.9321 | **1.0** | 0.9989 | 0.9243 |
| cm4-optics01 | 24.1075 | 0.64 | 0.9264 | **1.0** | 0.9935 | 0.8280 |
| cm4-agglomerative | 15.6945 | 0.03 | 0.9328 | 0.9303 | 0.9996 | 0.9510 |
| cm4-localoutlier04 | 6.5250 | 0.05 | 0.9326 | 0.9306 | 0.9994 | 0.9433 |
| cm4-birch01 | 0.7305 | **0.02** | **0.9330** | 0.9304 | **0.9997** | **0.9587** |
| cm4-kmeans | 0.2084 | 26.26 | 0.7119 | 0.6680 | 0.7373 | 0.5911 |

*F-M*: Fowlkes–Mallows index; *ACC*: unsupervised clustering accuracy; **nearestneighbors02:** n_neighbors = 5; **dbscan01:** eps = 3.25, min_samples = 5; **kmeans:** n_clusters = 2; **birch01:** n_clusters = 2, threshold = 0.5; **optics01:** min_cluster_size = int(len(data)*0.5); **agglomerative:** n_clusters = 2, affinity = 'euclidean', linkage = 'complete'; **localoutlier04:** n_neighbors = int(len(data)*0.1), contamination = 0.0005 from our corpus and the global model is *wiki-news-300d-1M.vec*.

We define a distance from redundant word $w_j$ to the quad set $Q$ as,

$$dist(w_j, Q) = \frac{\sum_{k}^{Q} sim_{[0,1]}(w_j, q_k)}{|Q|} \quad (26)$$

where $|Q|$ is size of the quad set, and $sim_{[0,1]}$ is the normalized similarity with a range from 0 to 1.

From (26), the sum distance between set $N$ (redundant words) and set $Q$ (quad items) for labeled sentence $ls_i$ is:

$$sum(ls_i) = \sum_{j}^{N} dist(w_j, Q) \quad (27)$$

Similarity, the product distance between set N and set Q for labeled sentence $ls_i$ is:

$$prod(ls_i) = -\log\left(\prod_{j}^{N} \log\left(\frac{1}{dist(w_j, Q)}\right)\right) \quad (28)$$

The key purpose here is to check the relatedness (or the similarity from local/global word2vec models) between redundant words and the quad items to decide which labeled sentences or even words we consider as noises and should take out of the corpus. Our intuitive is the labeled sentence which has a low value of sum and product distances likely be kept rather than the ones have high value.

According to (26), if combining with TF, IDF, or other combinations, we can have more variant formulas. This combination even allows us to have about 30 different values. For example, the formula combining with TF is:

$$dist(w_j, Q) = \frac{tf(w_j) \times \sum_{k}^{Q} sim_{[0,1]}(w_j, q_k)}{|Q|} \quad (29)$$

Also, when using (29), there may be some variants of (28) where we must plus 1 to avoid the problem of zero logarithms. To calculate the sum and product distances of a labeled sentence by TF or IDF, we apply the same formulas as (27) and (28) but much simpler. The formulas now are just the sum of TF or IDF values and the logarithm of the product of these values.

Here is for the step of noise filtering. The baseline method is to rank labeled sentences by the number of their redundant words, and then pick the proper threshold (the maximum number of redundant words) when observing on the plot. Figure 6 shows the baseline method of noise filtering over Wikipedia properties. We can observe when the number of redundant words from 15, the cumulative rate of all datasets is higher 80%. However, this method does not indicate the relatedness between the redundant words and translation contexts. For example, *P26* represents the marriage relationship between two individuals, we thus prefer to keep redundant words that be fitted in this context rather than remove all of them.

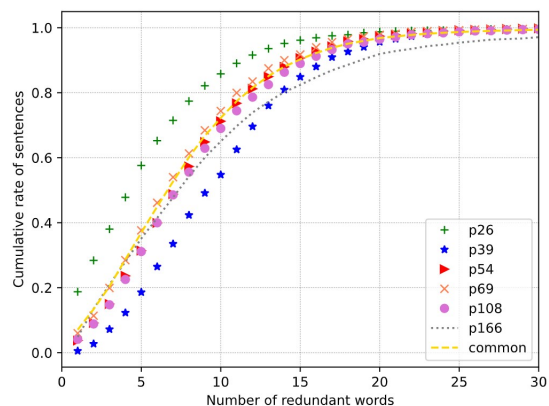

Fig. 6. The number of redundant words of labeled sentences by Wikipedia properties' datasets.

A better method is to use clustering algorithms to detect and remove noises (or outliers) based on the relatedness of redundant words in labeled sentences. The number of noises is also to be considered. If we remove a large number of noises, it may have an impact on the later translation methods in a way better but somehow, we also lost the naturality of the dataset. This entails the problem that we can not prove how well those translation methods are against the raw data. Our intention is to remove noises that indeed need to be removed, so the rate of noises removed may be relatively small. Therefore, we focus on density-based clustering algorithms. Lastly, the clustering result will be split into two sets: noise set, and noise-free set.

We convert labeled sentences to a set of vectors as the formula (28). Labeled sentence $ls_i$ is vector $v$ in form of $(prod_{tf}, prod_{idf}, prod_{local}, prod_{global})$. Then, we cluster these sentence vectors by four methods:
- *cm1*. clustering first, dimensionality reduction later.
- *cm2*. dimensionality reduction first, clustering later.
- *cm3*. hidden layer's data of vanilla autoencoder [34] uses for clustering first, dimensionality reduction later.
- *cm4*. hidden layer's data of denoising autoencoder uses for clustering first, dimensionality reduction later.

We also take *Candidates* as the ground truth and use this set to measure the clustering quality by these metrics:
- Fowlkes–Mallows index (FM): is used to measure the similarity between two clusters, ranging from 0 to 1 [35]. A higher value of FM shows a greater similarity between the clusters.
- Unsupervised clustering accuracy (ACC): is a metric to calculate the matching between clusters and the ground truth [36], also is the range of 0 and 1. The clustering performance is better when the value of ACC is higher [37].
- Purity: This metric is used for evaluating the simplicity and transparency of the data clustering [38]. Any value is close to 0 indicates a bad clustering while a good clustering has a value near 1 [39].
- Silhouette index: This metric represents the consistency of the clustering data, ranging from -1 to +1. It measures the similarity of a object to other objects in the same cluster and in other different clusters. A high value shows the proper match to clusters and a low value indicated the poor match to clusters [40, 41].

Table 12 presents the comparisons of prominent clustering algorithms by various metrics. In terms of the executed time, Nearest Neighbors, K-means, BIRCH, and DB-SCAN are relatively fast compared to OPTICS and Agglomerative. Observing all methods, we can see that the rate of noise is inversely proportional to quality metrics when reducing the number of data points in the noise-free cluster may reduce the compactness, similarity or consistency between clusters. K-means is also the only method that has metrics with a low value due to a high rate of noises produced while other algorithms obtain this rate of less than 0.1%.

Method *cm4-birch01* is one of the methods which adapts our requirements with a minimum rate of noises and high metrics. In contrast, method *cm3-kmeans* is the worst performance. There is no much difference of FM, ACC, Purity and Silhouette values in these algorithms: Nearest Neighbors, Agglomerative clustering, Local Outlier Factor, Birch, and DBSCAN (except the Silhouette value in method *cm1-dbscan01*).

The three methods *cm2*, *cm3*, and *cm4* is better than *cm1* when they indicate the faster executed time, the least rate of noise and higher metric values in all clustering algorithms. It is not easy to determine the better between *cm2* and *cm3* when we have to depend on specific cases. The executed time of method *cm4* is slower than *cm2* and *cm3* but other metrics are slightly better. This can be explained when swapping noise data for the denoising process, clustering algorithms probably take more time to fit the new noise data into clusters.

Overall, all clustering algorithms should go with method *cm2* because this method offers the simple enough for the clustering instead of training through autoencoder networks. Note that we exclude the executed time of training in autoencoder networks in *cm3* and *cm4* in the comparions in Table 12. For the optimization cases, *cm3* and *cm4* are recommended while *cm1* is still the optional choice for some situations.

*6.4. Relationships between sentence predicates and Wikidata properties or qualifiers*

In this section, we will perform the experiment for the predicate matching as Section 4.8 but in a different way in order to observe relationships between the sentence predicates against Wikidata properties and qualifiers. The example in Section 6.2 will be reused for the illustration more understanding.

Raw sentence:
  She and David Birney <u>divorced</u> in 1989.
Labeled sentence (also the candidate sentence):
  [s] and [o0] <u>divorced</u> in [o0:P582-qualifier].
Quad $T$ (a quad result when mapping to Wikidata):
  (Q272022, P26, Q1173755, P582)

In the example, *P26* is a Wikidata property shown the marriage relationship between two people. *Q272022* is a Wikidata item described for American actress Meredith Baxter, who divorced American actor David Birney (*Q1173755*) in 1989 (*P582* is the end time of a marriage relationship).

Given quad $T_i$ (*s*, *p*, $O_n$, $Q_m$) whose *s* is the subject, *p* is the predicate, $O_n$ is set of objects, and $Q_m$ is a set of qualifiers of an labeled sentence. Table 13 describes these alignments and we have only one object and one qualifier in the example sentence. Now, our task is to compare the sentence predicate *p* (*divorced*) to its Wikidata property (*P26*) and quad qualifiers (*P582*). We also deloy two Word2vec models, local model and global model as Section 6.3.

Table 13

The alignments between quad structure, labeled sentence's items, and Wikidata of the example sentence.

| Quad items | Labeled items | Wikidata items or properties |
|---|---|---|
| s | [s] | Q272022 |
| p | divorced | P26 |
| $O_n$ | [o0] | Q1173755 |
| $Q_m$ | [o0:P582-qualifier] | P582 |

In the local mode, base on (26), we define the local distance $dist_{local}$ between predicate *p* and qualifiers *Q* of the labeled sentence is:

$$dist_{local}(p,Q) = \frac{\sum_i^Q sim(p,q_i)}{|Q|} \quad (30)$$

which *Q* is a set of labeled qualifiers such as *{'[o0:P582-qualifier]'}* in the example since we have used the labeled sentences as the input of word embedding training. We do not need to measure the distances between the predicate p to other labeled items such as *[s]* or *[o0]*. All the labeled sentences contain them so these distances likely are approximately each other and have less contribute much to figure out the relationship of results.

In the global model, we manually create a dictionary containing terms for each Wikidata property (*P26*). In other words, each Wikidata property has its own set of definition terms. We define distance $dist_{global}$ from predicate *p* to Wikidata property *Wp* as:

$$dist_{global}(p,Wp) = \frac{\sum_i^T sim(p,t_i)}{|T|} \quad (31)$$

with *T* is a definition term set of Wikidata property *Wp*. For example, the definition term set of *P26* in our corpus is *{'spouse', 'wife', 'married', 'marry', 'marriage', 'partner', 'wedded', 'wed', 'wives', 'husbands', 'spouses', 'husband'}*.

Now, we define the revelant score *rs* in the normalized range from 0 to 1 between predicate *p* and Wikidata property *Wp* is:

$$rs_{[0,1]}(p,Wp) = \log\left(\frac{3(dist_{local} \times dist_{global} \times tf(p))}{dist_{local} + dist_{global} + tf(p)}\right) \quad (32)$$

where *tf(p)* is the TF value of predicate *p* in the corpus of labeled sentences.

Table 14

The top 10 predicates and their relevant scores by various Wikidata properties.

| Wikidata properties | Sentence predicates & Relevant scores |
|---|---|
| P26 | (**married**, 1), (met, 0.27), (divorced, 0.23), (became, 0.22), (began, 0.21), (announced, 0.19), (remarried, 0.18), (born, 0.18), (appeared, 0.17), (left, 0.17) |
| P39 | (**elected**, 1), (served, 0.63), (became, 0.60), (appointed, 0.59), (resigned, 0.43), (ran, 0.39), (named, 0.39), (succeeded, 0.39), (elevated, 0.38), (remained, 0.37) |
| P54 | (**signed**, 1), (joined, 0.89), (made, 0.71), (moved, 0.67), (returned, 0.59), (scored, 0.57), (played, 0.57), (left, 0.56), (loaned, 0.52), (transferred, 0.49) |
| P69 | (**graduated**, 1), (received, 0.69), (earned, 0.55), (attended, 0.52), (completed, 0.4), (studied, 0.39), (entered, 0.35), (obtained, 0.33), (went, 0.32), (enrolled, 0.31) |
| P108 | (**joined**, 1), (became, 0.79), (appointed, 0.77), (left, 0.73), (worked, 0.73), (retired, 0.68), (began, 0.65), (moved, 0.62), (started, 0.61), (returned, 0.60) |
| P166 | (**awarded**, 1), (received, 0.88), (won, 0.78), (elected, 0.55), (appointed, 0.51), (named, 0.43), (made, 0.43), (shared, 0.41), (became, 0.39), (nominated, 0.35) |

Table 14 shows the relationships between top 10 sentence predicates and their Wikidata properties ranked by the relevant scores. These scores are normalized to the range [0, 1] so the highest value will always be 1. There are some reasonable verbs going with Wikidata property *P26* (the marriage relationship) such as "*married*", "*divorced*", and "*remarried*". However, in another cases such as "*born*" or "*appeared*", we can see that there are no any clear connection to a marriage relationship. The relationships also happen between sentence predicates against Wikidata properties and qualifiers. Table 15 lists these relationships on two qualifiers, *P580* (start time) and *P582* (end time). For example, a player career in a club (*P54*) can start (*P580*) with predicate "*joined*" and end with (*P580*) predicate "*left*".

Table 15

The relationship between Wikidata properties and qualifiers and sentence predicates over the qualifier *P580* (start time) and *P582* (end time).

| Wikidata properties & qualifiers | Sentence predicates |
|---|---|
| P26 – P580 | **married**, met, became, began, announced |
| P26 – P582 | **divorced**, married, separated, died, continued |
| P39 – P580 | **elected**, appointed, became, served, elevated |
| P39 – P582 | **resigned**, served, elected, remained, retired |
| P54 – P580 | signed, **joined**, made, moved, returned |
| P54 – P582 | **left**, signed, released, played, scored |
| P69 – P580 | enterer, **enrolled**, began, admitted, matriculated |
| P69 – P582 | **graduated**, received, earned, attended, completed |
| P108 – P580 | **joined**, became, appointed, began, worked |
| P108 – P582 | **left**, retired, worked, resigned, served |
| P166 – P580 | **appointed**, elected, named, awarded |
| P166 – P582 | **participated** |

In Table 14 and Table 15, most of predicates are verbs in the past tense because two reasons. First, we do not apply the lemmatization process in order to keep the naturality of dataset at best for the future translation task. Second, according to Table 8, most of our qualifiers are Wikidata properties with datatype "time" so the events in mapped sentences usually happen in the past.

These relationships usually happen on a single property or a pair of property-qualifier, but not limit to a set of properties/qualifiers against sentence verbs. From that, we can apply an attention mechanism in neural networks to preload the candidate texts for improving the translation performance [42, 43, 44, 45, 49]. In another method, we can use Fill-Mask and text generation models such as BERT [51] or Roberta [52] to fill in the content of *[MASK]* tags or generate the next text from inputs or just to validate the translation performance.

*6.5. Selected examples*

Table 16 lists some examples taken from our corpus by order: the quad, the raw sentence, the labeled sentence, their values of IF/IDF, and their local and global distances as (27, 28). All the metrics have the tendency of proportional to the number of redundant words. The lower values indicate better results.

The sentence main verbs are in bold, some of which can be found in Table 15 in relationships between them with Wikidata properties and qualifiers. Some words/terms are in double underline because they fit the statement context. This hints us to extend the mapping ability by combining more triples and quadruples on Wikidata or even shows us the missing mapping in our experimental data.

# 7. Conclusion

This paper presents our method in the mapping process from Wikidata statements, especially ones with qualifiers, to single Wikipedia sentences. This process is split into sub-processes: subject matching, predicate matching, object matching, qualifier matching, and external matching. The input is a set of triples and quadruples built from Wikidata statements, which are organized by different classifications. Wikipedia mapped sentences and their labeled versions are what we will receive in the output.

We also describe some techniques for content selection and data collection to be sure that we can gather the data effectively by our mapping algorithms. For the evaluation, entity linking methods are used to evaluate sub-mapping processes independently. We

Table 16

Some selected examples from the corpus.

| Quad, Mapped raw sentence, and Labeled sentence | Redundant words, TF, IDF, Local and Global distances | | | | |
|---|---|---|---|---|---|
| Q123849, **P26**, Q22910017, P580<br>In 1981, Seymour married David Flynn.<br>In [o0:P580-qualifier] , [s] **married** [o0] . | 1 | 2.21 | 1.30 | 2.79 | 0.90 |
| Q5590131, **P26**, Q16200324, P580<br>Gowri Pandit married Bollywood actor Nikhil Dwivedi on 7 March 2011.<br>[s] **married** Bollywood [o0:P106-occupation] [o0] on [o0:P580-qualifier] . | 2 | 3.33 | 3.19 | 5.62 | 1.30 |
| Q6152775, **P26**, Q2271796, P580<br>On October 25, 1997, she married now-NFL Commissioner Roger Goodell and resides in Westchester, New York, with their twin daughters, born in 2001.<br>On [o0:P580-qualifier] , [s] **married** now - <u>NFL Commissioner</u> [o0] and <u>resides</u> in <u>Westchester , New York</u> , with their twin <u>daughters</u> , <u>born</u> in <u>2001</u> . | 12 | 16.57 | 23.21 | 25.29 | 8.78 |
| Q77156, **P39**, Q29034484, P580<br>In June 1920, Fehrenbach became Chancellor of Germany.<br>In [o0:P580-qualifier] , [s] **became** [o0] of [o0:P1001-applies_to_jurisdiction] . | 1 | 2.00 | 1.68 | 2.56 | 0.72 |
| Q8273282, **P39**, Q27169, P582<br>He left the party, but continued to sit as an independent MEP until 2002.<br>[s] **left** [det:the] <u>party</u> , but continued to sit as [det:a-an] <u>independent</u> [o0] until [o0:P582-qualifier] . | 6 | 9.82 | 10.62 | 15.41 | 4.69 |
| Q529647, **P54**, Q196107, P580<br>Tornaghi signed with Vancouver Whitecaps FC on 18 February 2014.<br>[s] **signed** with [o0] on [o0:P580-qualifier] . | 1 | 2.12 | 1.49 | 2.63 | 0.81 |
| Q217760, **P54**, Q221525, P580<br>On 30 January 2010, Wiltord signed with Metz until the end of the season.<br>On [o0:P580-qualifier] , [s] **signed** with [o0] until [det:the] end of [det:the] <u>season</u> . | 3 | 6.07 | 4.29 | 6.59 | 2.59 |
| Q703956, **P54**, Q8455, P1351-P582<br>He ended his career in 2007 with Pro Vercelli in Serie C2, scoring 10 goals.<br>[s] **ended** [s:poss] <u>career</u> in [o0:P582-qualifier] with [o0] in <u>Serie C2</u> , <u>scoring</u> [o0:P1351-qualifier] <u>goals</u> . | 6 | 10.65 | 10.36 | 16.18 | 5.11 |
| Q42728925, **P69**, Q16952, P582<br>Yan earned a bachelor's degree from Peking University in 1993.<br>[s] **earned** [det:a-an] <u>bachelor 's degree</u> from [o0] in [o0:P582-qualifier] . | 3 | 5.72 | 4.85 | 6.47 | 2.22 |
| Q48472222, **P108**, Q189022, P580<br>Green joined Imperial College London in 2016.<br>[s] **joined** [o0] in [o0:P580-qualifier] . | 1 | 2.11 | 1.51 | 2.55 | 0.85 |
| Q463124, **P108**, Q208046, P580-P582<br>Sullenberger was employed by US Airways and its predecessor airlines from 1980 until 2010.<br>[s] was **employed** by [o0] and its <u>predecessor airlines</u> [o0:P580-qualifier] until [o0:P582-qualifier] . | 4 | 5.01 | 7.54 | 8.77 | 3.01 |
| Q2850193, **P166**, Q1537127, P585<br>Wall won the Eugene O'Neill Award in 2008.<br>[s] **won** [det:the] [o0] in [o0:P585-qualifier] . | 1 | 1.99 | 1.70 | 2.36 | 0.89 |

realize that the very rare chance to have a mapping successfully when we receive 18510 sentences over 113913 articles scanned. The reason may come from the language diversity or our narrow working scope on single sentences. Therefore, we may face the problem of low resources in the translation task.

We perform several statistical methods on various datasets, including labeled candidate sentences extracted from the output by our definition of redundant words. We consider these candidates as the ground-truth set in comparison to other datasets. Some density-based clustering algorithms are applied for noise filtering. We recommend Nearest Neighbors, BIRCH, and DB-SCAN, in particular for the quick executed time and the ability to control the rate of noises as well as the method of dimensionality reduction before the clustering. We prefer to have the rate of filtered noises at least as possible to keep the naturality of the dataset. However, it should depend on the input requirements by cases. Based on word embedding models, we also search for relationships between sentence components and Wikidata properties/qualifiers. These relationships are indeed helpful for the attention mechanism of the translation process.

In the future, we will generate clauses or short passages by applying various methods such as Beam Search, deep neural networks (Seq2Seq, LSTM, CNN, etc), and deep neural graphs. We also apply word embedding models such as BERT embeddings [46] and universal sentence encoder [47] to work with semantic similarity on phrases or short texts.

### Acknowledgements

The work was done with partial support from the Mexican Government through the grant A1-S-47854 of the CONACYT, Mexico and grants 20211784, 20211884, and 20211178 of the Secretaría de Investigación y Posgrado of the Instituto Politécnico Nacional, Mexico.